\documentclass[letterpaper]{article} % DO NOT CHANGE THIS
\usepackage{aaai24}  % DO NOT CHANGE THIS
\usepackage{times}  % DO NOT CHANGE THIS
\usepackage{helvet}  % DO NOT CHANGE THIS
\usepackage{courier}  % DO NOT CHANGE THIS
\usepackage[hyphens]{url}  % DO NOT CHANGE THIS
\usepackage{graphicx} % DO NOT CHANGE THIS
\usepackage{natbib}  % DO NOT CHANGE THIS AND DO NOT ADD ANY OPTIONS TO IT
\usepackage{caption} % DO NOT CHANGE THIS AND DO NOT ADD ANY OPTIONS TO IT
\usepackage{algorithm}
\usepackage{algorithmic}
\usepackage{newfloat}
\usepackage{listings}
\DeclareCaptionStyle{ruled}{labelfont=normalfont,labelsep=colon,strut=off} % DO NOT CHANGE THIS
\floatstyle{ruled}
\newfloat{listing}{tb}{lst}{}
\floatname{listing}{Listing}
\usepackage{amsmath}
\usepackage{amssymb}

\title{S2WAT: Image Style Transfer via Hierarchical Vision Transformer \\ Using Strips Window Attention}
\author {
    Chiyu Zhang\textsuperscript{\rm 1,\rm 2},
    Xiaogang Xu\textsuperscript{\rm 3,4,}\thanks{Corresponding authors},
    Lei Wang\textsuperscript{\rm 1},
    Zaiyan Dai\textsuperscript{\rm 1},
    Jun Yang\textsuperscript{\rm 1,5,*}
}
\affiliations {
    \textsuperscript{\rm 1}Sichuan Normal University, 
    \textsuperscript{\rm 2}Nanjing University of Aeronautics and Astronautics, 
    \textsuperscript{\rm 3}Zhejiang Lab,
    \textsuperscript{\rm 4}Zhejiang University\\
    \textsuperscript{\rm 5}Visual Computing and Virtual Reality Key Laboratory of Sichuan Provience\\
    \{alienzhang19961005, xiaogangxu00, londmi9, zaiyan.dai\}@gmail.com, jkxy\_yjun@sicnu.edu.cn
}

\usepackage{bibentry}
\begin{document}

\maketitle

%%%%%%%%% ABSTRACT
\begin{abstract}
Transformer's recent integration into style transfer leverages its proficiency in establishing long-range dependencies, albeit at the expense of attenuated local modeling.
This paper introduces Strips Window Attention Transformer (S2WAT), a novel hierarchical vision transformer designed for style transfer. S2WAT employs attention computation in diverse window shapes to capture both short- and long-range dependencies. The merged dependencies utilize the ``Attn Merge" strategy, which adaptively determines spatial weights based on their relevance to the target.
Extensive experiments on representative datasets show the proposed method's effectiveness compared to state-of-the-art (SOTA) transformer-based and other approaches. The code and pre-trained models are available at \url{https://github.com/AlienZhang1996/S2WAT}.
%%this approach has drawbacks in ignoring local modeling.
%and large computation cost. 
%%%%%In this paper, we propose a novel hierarchical vision transformer for style transfer, called Strips Window Attention Transformer (S2WAT). S2WAT conducts attention computation within various shapes of windows, capturing sufficient short- and long-range dependencies. The extracted dependencies are then merged by our newly designed ``Attn Merge" strategy, where spatial weights of dependencies are adaptively decided by their importance to the target. 
%%By choosing suitable window shapes, the efficiency of S2WAT can be guaranteed theoretically. 
%%Extensive qualitative and quantitative 
%%%Extensive experiments are conducted on representative datasets, demonstrating the effectiveness of the proposed method, compared with state-of-the-art transformer-based and other approaches. The code can be found in the supplementary material.

\end{abstract}

% The code can be found in the supplementary material.
% The code and models are available at \url{https://github.com/AlienZhang1996/S2WAT}.

%%%%%%%%% BODY TEXT
\section{Introduction}

{\bf Background}. Image style transfer imparts artistic characteristics from a style image to a content image, evolving from traditional~\cite{p1} to iterative~\cite{p3,p4} and feed-forward methods~\cite{p11,p17}. Handling multiple styles concurrently remains a challenge, addressed by Universal Style Transfer (UST)~\cite{p37,p94,p96}. This sparks innovative approaches like attention mechanisms for feature stylization~\cite{p38,p39,p42}, the Flow-based method~\cite{p43} for content leakage, and Stable Diffusion Models (SDM) for creative outcomes \cite{p100}. New neural architectures, notably the transformer, show remarkable potential. \cite{p45} introduces StyTr2, leveraging transformers for SOTA performance. However, StyTr2's encoder risks losing information due to one-time downsampling, impacting local details with global MSA (multi-head self-attention).

%%%%Image style transfer aims at transferring the artistic characteristics from a style image to a content one. It has experienced technological updates from traditional methods \cite{p1}, to iterative approaches \cite{p3,p4}, and to feed-forward algorithms \cite{p11,p17}. However, these methods can not generalize to handle multiple types of styles simultaneously. Universal Style Transfer (UST) \cite{p37,p94,p96} is further designed to deal with such a challenge. Afterwards, attention mechanism has been concerned \cite{p38,p39,p42} as a new way to stylize features, the Flow-based \cite{p43} method is designed to solve the content leak problem and \cite{p100} introduces stable diffusion models (SDM) to receive creative results. Recently, \cite{p45} propose StyTr2 which makes another step forward by utilizing the ability of transformer to capture long-range dependencies. However, the encoder of StyTr2 may drop the useful information by the one-time downsampling and the local information is not well modeled due to the global MSA (multi-head self-attention).\label{c1}

\begin{figure}[t]
\centerline{\includegraphics[width=1.0\linewidth]{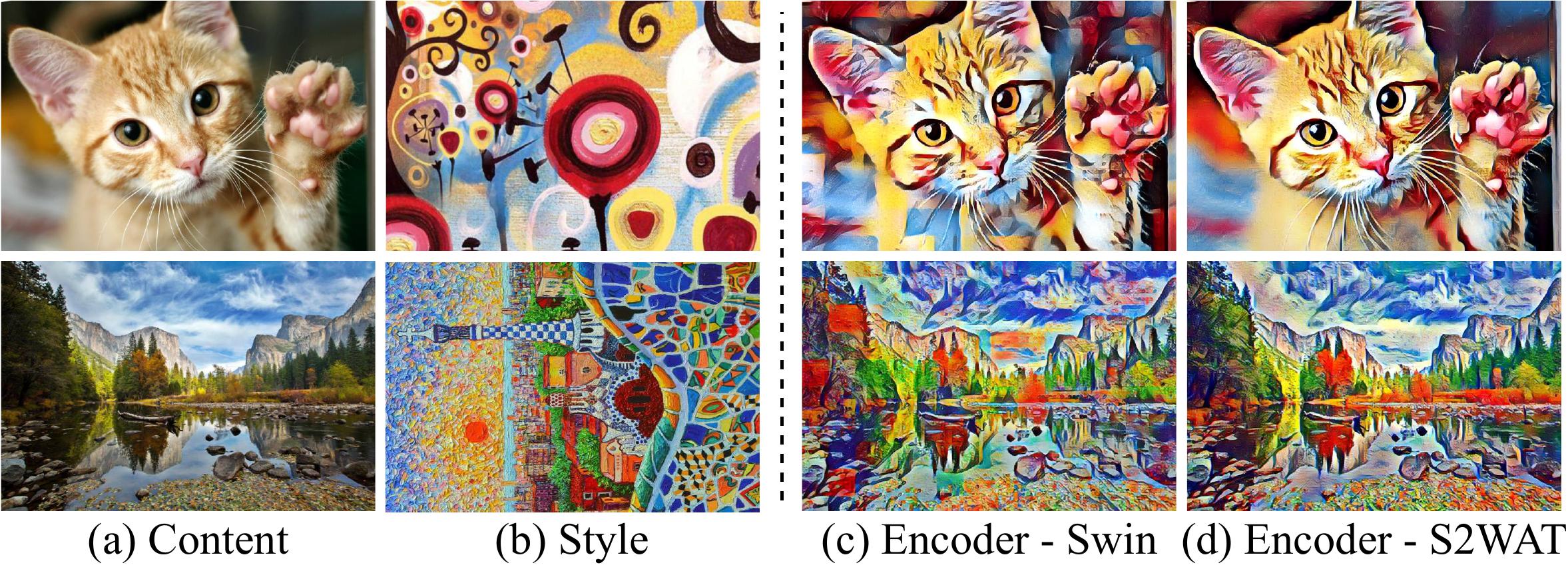}}
% \vspace{-0.1in}
\caption{Illustration for locality problem. (c) Results of the Swin-based model. (d) Results from our S2WAT.}
\label{fig2}
% \vspace{-1.8em}
\end{figure}

% \noindent\textbf{Challenge.}
{\bf Challenge}. To enhance the transformer's local modeling capability, recent advancements propose the use of window-based attention computation, exemplified by hierarchical structures like Swin-Transformer~\cite{p62}. However, applying window-based transformers directly for feature extraction in style transfer can lead to grid-like patterns, as depicted in Fig.~\ref{fig2} (c). This arises due to the localized nature of window attention, termed the \textit{locality problem}. While window shift can capture long-range dependencies~\cite{p62}, it necessitates deep layer stacks, introducing substantial model complexity for style transfer, particularly with high-resolution samples.

{\bf Motivation and Technical Novelty}.Diverging from current transformer-based approaches, we introduce a novel hierarchical transformer framework for image style transfer, referred to as S2WAT (\textbf{S}trip\textbf{s} \textbf{W}indow \textbf{A}ttention \textbf{T}ransformer). This structure meticulously captures both local and global feature extraction, inheriting the efficiency of window-based attention.
In detail, we introduce a distinct attention mechanism (\textbf{S}tri\textbf{p}s \textbf{W}indow Attention, SpW Attention) that amalgamates outputs from multiple window attentions of varying shapes. These diverse window shapes enhance the equilibrium between modeling short- and long-range dependencies, and their integration is facilitated through our devised ``Attn Merge" technique.

%%%%Different from existing transformer-based methods, we propose a new hierarchical transformer structure for image style transfer, called S2WAT (\textbf{S}trip\textbf{s} \textbf{W}indow \textbf{A}ttention \textbf{T}ransformer), explicitly modeling both local and global feature extraction.
%%%%Specifically, we present a new type of attention mechanism (\textbf{S}tri\textbf{p}s \textbf{W}indow Attention, SpW Attention) whose outputs are merged from multiple window attentions with various window shapes. These different window shapes allow a better balance between short- and long-range dependencies modeling, and they can be merged via our designed ``Attn Merge".

In this paper, we formulate the SpW Attention in a simple while effective compound mode, which encompasses three window types: horizontal strip-like, vertical strip-like, and square windows.
The attention computations derived from strip windows emphasize long-range modeling for extracting non-local features, while the square window attention focuses on short-range modeling for capturing local features.

%%%In this paper, we formulate the SpW attentuon with a horizontal strip-like, a vertical strip-like, and a square window for ``SpW Attention". 
%%%Attention computations from the strip windows response for the long-range modeling to extract non-local features, and the square window attention can implement short-range modeling to obtain local features.
%%Moreover, compared with global MSA and window-based MSA respectively, our designed strips-window-based MSA has the advantage in computation cost which can be theoretically explained or long-range dependencies modeling.
% Moreover, compared with global MSA and window-based MSA, our designed strips-window-based MSA has the advantage in computation cost which can be theoretically explained.
% The ``Attn Merge" takes the attention outputs from different windows, compute the spatial correlation between them. The computed correlation scores can be served as the weights to merge. Compared with other static merge technique, e.g., sum and average, ``Attn Merge" can adaptively decide the importance of different window attentions.

%%%Another significant question is how to merge dependencies from multiple windows.
%%%We propose ``Attn Merge" to solve this.
Furthermore, the ``Attn Merge" method combines attention outputs from various windows by computing spatial correlations between them and the input. These calculated correlation scores serve as merging weights. In contrast to static merge strategies like summation and concatenation, ``Attn Merge" dynamically determines the significance of different window attentions, thus enhancing transfer effect.

{\bf Contributions}. Extensive quantitative and qualitative experiments are conducted to prove the effectiveness of the proposed framework, including a large-scale user study.
The main contributions of our work include:
\begin{itemize}

\item We introduce a pioneering image style transfer framework, S2WAT, founded on a hierarchical transformer. This framework adeptly undertakes both short- and long-range modeling concurrently, effectively mitigating the challenge of locality issues.
%%%We propose a novel image style transfer framework based on a hierarchical transformer, called S2WAT, which can simultaneously conduct both short- and long-range modeling to avoid the locality problem.
%%has multi-scaled intermediate representation and eliminates the locality problem.

\item %%We propose a new type of feature fusion scheme named Attn Merge which is capable of merging results from multiple attentions of arbitrary window size. And SpW Attention is the attention when window sizes are set to a horizontal strip-like, a vertical strip-like, and a square one.
% We design a new type of attention computation in the transformer, called SpW Attention, which can adaptively merge the attention outputs from various types of window attentions.
We devise a novel attention computation within the transformer for style transfer, termed SpW Attention. This mechanism intelligently merges outputs from diverse window attentions using the ``Attn Merge" approach.

\item We extensively evaluate our proposed S2WAT on well-established public datasets, demonstrating its state-of-the-art performance for the style transfer task.
%%We conduct extensive experiments on public representative datasets to prove the SOTA performance of our proposed S2WAT.
\end{itemize}

% \vspace{-1.0em}

\section{Related Work}

\noindent\textbf{Image Style Transfer.}
Style transfer methods can fall into single-style~\cite{p14}, multiple-style~\cite{p17}, and arbitrary-style (UST)~\cite{p46,p94,p95} categories based on their generalization capabilities. Besides models based on CNNs, recent works include Flow-based ArtFlow~\cite{p43}, transformer-based StyTr2~\cite{p45}, and SDM-based InST~\cite{p100}. ArtFlow, with Projection Flow Networks (PFN), achieves content-unbiased results, while IEST~\cite{p42} and CAST~\cite{p46}  use contrastive learning for appealing effects. InST achieves creativity through SDM. Models like \cite{p44,p93,p101} use transformers to fuse image features, and \cite{p91,p92} encode text prompts for text-driven style transfer. StyTr2 leverages transformers as the backbone for pleasing outcomes. Yet, hierarchical transformers remain unexplored in style transfer.
\noindent\textbf{Hierarchical Vison Transformer.}
% Recent interest in hierarchical transformer architectures, exemplified by LeViT \cite{p58}, CvT \cite{p89}, PVT \cite{p59}, and MViT \cite{p60}, emphasis a focus on global information but ignores local modeling. Despite effective local information capture by Swin through shifted windows, it encounters the locality problem in style transfer (see Fig.~\ref{fig2}). Attempts like global MSA insertion or Mix-FFN \cite{p97} convolution prove ineffective (Pre-Analysis section, and appendix). In the realm of style transfer, a promising direction involves advancing a new transformer architecture that encompasses both short- and long-range dependencies while mitigating the locality problem.
Lately, there has been a resurgence of interest in hierarchical architectures within the realm of transformers. Examples include LeViT \cite{p58} \& CvT \cite{p89}, which employ global MSA; PVT \cite{p59} \& MViT \cite{p60}, which compress the resolution of K \& V. However, in these approaches, local information is not adequately modeled. 
While Swin effectively captures local information through shifted windows, it still gives rise to the locality problem when applied to style transfer (see Fig.~\ref{fig2}).
Intuitive attempts, such as inserting global MSA (see Section Pre-Analysis) or introducing Mix-FFN \cite{p97} by convolutions (see appendix), are powerless for locality problem.
In the context of style transfer, a promising avenue involves advancing further with a new transformer architecture that encompasses both short- and long-range dependency awareness and possesses the capability to mitigate the locality problem.

%%%%%%Recently, the hierarchical architecture has been brought back to transformers. LeViT \cite{p58} \& CvT \cite{p89} still use global MSA and PVT \cite{p59} \& MViT \cite{p60} compress the resolution of K \& V, where the local information is not well modeled. 
%%Though Swin \cite{p62,p63} and CSWin \cite{p90} are able to capture local information with shifted windows or strip windows, Swin will cause the locality problem for style transfer (see Fig.~\ref{fig2})
%%while the concatenating operation (taken by CSWin) will have a bad effect on training in image style transfer. 
%%%%%%Though Swin \cite{p62,p63} is able to capture local information with shifted windows, Swin causes the locality problem for style transfer (see Fig.~\ref{fig2}).
% The strips window attention of CSWin \cite{p90} is similar to the proposed SpW Attention, but the fusion scheme of results from attentions is different. CSWin uses the concatenating operation while S2WAT utilizes the novel ``Attn Merge" which is significantly superior in image style transfer (see Fig.~\ref{fig11}, column 5). Differences from Longformer \cite{p98} and iLAT \cite{p99} are detailed in Appendix.
%%%%%For style transfer, it is promising to make another step forward through a new transformer with perceptivity of short- \& long-range dependencies and the ability to erase the locality problem.

% \vspace{1mm}
\noindent\textbf{Differences with Other Methods}
% The strips window attention of CSWin \cite{p90} is similar to the proposed SpW Attention, but the fusion scheme of results from attentions is different. CSWin uses the concatenating operation while S2WAT utilizes the novel ``Attn Merge" which is significantly superior in image style transfer (see Fig.~\ref{fig11}, column 5). Differences from Longformer \cite{p98} and iLAT \cite{p99} are detailed in Appendix.
While the attention mechanism in certain prior methods may share similarities with the proposed SpW Attention, several key distinctions exist. 1) The fusion strategy stands out: our proposed ``Attn Merge" demonstrates remarkable superiority in image style transfer. 2) In our approach, all three window shapes shift based on the computation point, and their sizes dynamically adapt to variations in input sizes.
Detailed differentiations from previous methods, such as CSWin, Longformer, and iLAT have been outlined in the Appendix.

%%%The Attention in some previous methods may resemble the proposed SpW Attention. But here are several main differences as follows: 1) the fusion strategy is distinctive; the proposed ``Attn Merge" is significantly superior in image style transfer (see Fig.~\ref{fig11}, column 5); 2) all of the three shapes of windows will move according to the computing point, and with different sizes of inputs the window size will be changed correspondingly. Detailed differences from previous methods, such as CSWin \cite{p90}, Longformer \cite{p98}, and iLAT\cite{p99}, are left in Appendix.

% \vspace{-0.4em}
\section{Pre-Analysis}\label{section: Pre-analysis}
% \vspace{-0.05in}

%%Before introducing the proposed S2WAT, we plan to make a pre-analysis to reveal why the grid-like results (locality problem) will be generated when applying Swin to style transfer directly. We think the reason is the locality of window attention. 
Our preliminary analysis aims to unveil the underlying causes of grid-like outcomes (locality problem) that arise when directly employing Swin for style transfer. Our hypothesis points towards the localized nature of window attention as the primary factor.
%%To verify the correctness of this opinion, we conduct experiments in four aspects as discussed in Section \ref{section: Location of Global MSA} – \ref{section: Attention Maps}.
% To validate this hypothesis, we undertake experiments across four distinct facets as discussed in Section \ref{section: Location of Global MSA} – \ref{section: Attention Maps}. The details of the models tested in this part can be found in Appendix.
To validate this hypothesis, we undertake experiments across four distinct facets as discussed in this Section. The details of the models tested in this part can be found in Appendix.

%%%To verify this opinion, we conduct experiments in four aspects as discussed in Section \ref{section: Location of Global MSA} – \ref{section: Attention Maps}. The details of models tested in this part can be found in Appendix.

% The models tested in this part adopt the encoder-transfer-decoder architecture, where the encoder is based on a Swin, which has 3 stages, and each stage has 2 successive Swin transformer blocks (we call it layer hereafter). When a layer is replaced by global MSA (multi-head self-attention), the window won't be shifted if it was. The transfer module and decoder are the same as those of S2WAT.

%%\subsection{Location of Global MSA}
% \vspace{-0.5em}
\subsection{Global MSA for Locality Problem}
\label{section: Location of Global MSA}
\begin{figure}[t]
\centerline{\includegraphics[width=1.0\linewidth]{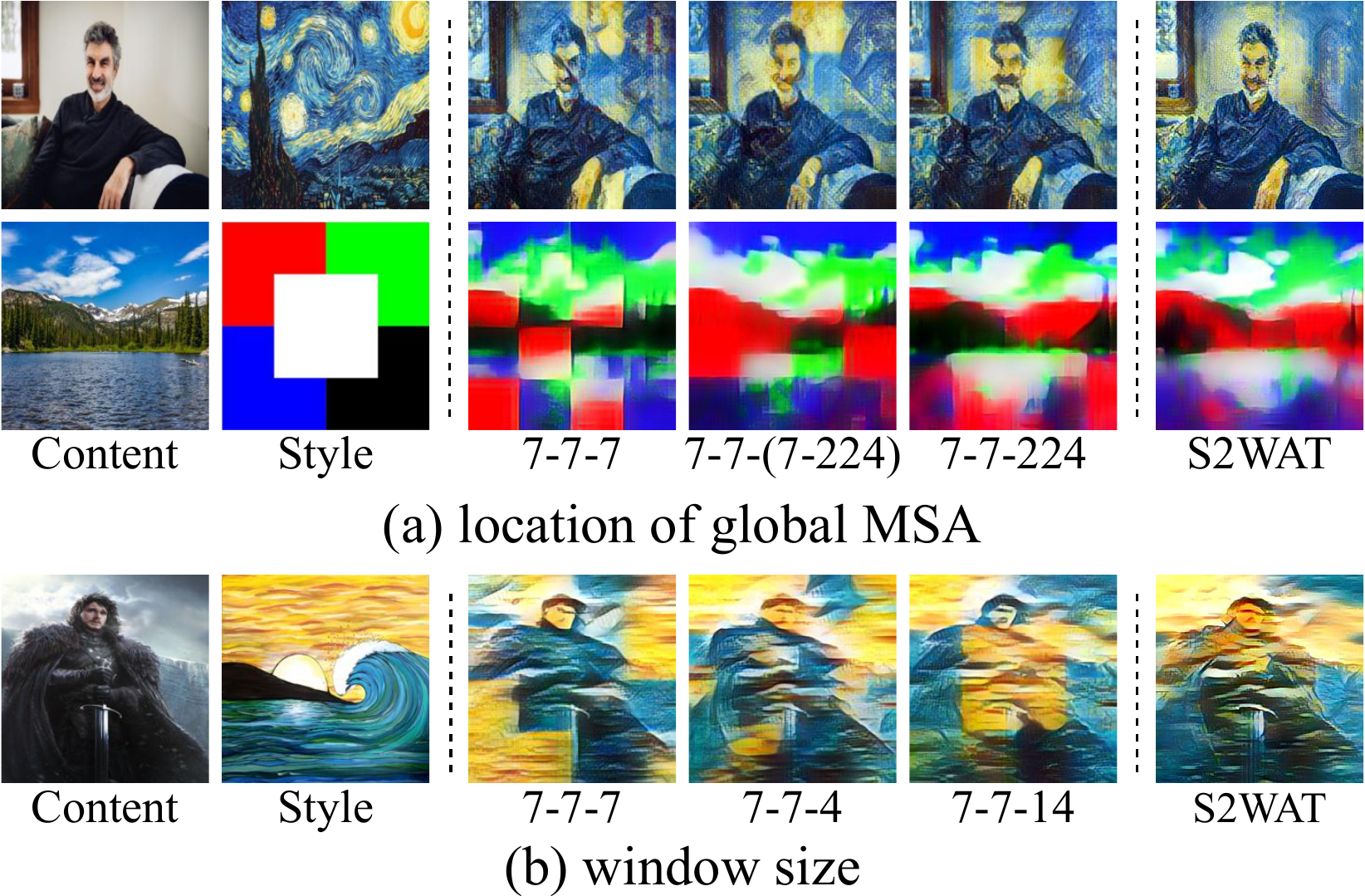}}
% \vspace{-0.1in}
\caption{Results of the Swin-based encoder experiments. \textit{7-7-7} means the Swin used has 3 stages (each stage with 2 layers) and 7 is the window size of each layer. \textit{7-7-(7-224)} denotes the window size of the last layer in the last stage is 224 which represents global MSA. (a) Results of experiments inserting global MSA in certain layers. (b) Results of experiments changing the window size.}
\label{fig3}
% \vspace{-1.5em}
% \vspace{-0.18in}
\end{figure}

% \vspace{-0.5em}
The locality problem should be relieved or excluded when applying global MSA instead of window or shifted window attention, if the locality of window attention is the culprit. 
In the Swin-based encoder, we substitute the last one or two window attentions with global MSA, configuring the window size for target layers at 224 (matching input resolution). Fig.~\ref{fig3} (a) presents the experiment results, highlighting grid-like textures at a window size of 7 (column 3) and block-like formations when the last window attention is swapped with global MSA (column 4). While replacing the last two window attentions with global MSA effectively alleviates grid-like textures, complete exclusion remains a challenge.
%%We replace the last one or two window attentions in the Swin-based encoder with global MSA by setting the window size of the target layers to 224 (input resolution is 224). Fig.~\ref{fig3} (a) shows the results of experiments. Apparent grid-like textures appear when the window size is all 7 (column 3). Blocks cloud also be seen when the last window attention is replaced by global MSA (column 4). Further, when we substitute the last two window attentions with global MSA, grid-like textures are effectively relieved but can not be excluded.
%%at some times (column 5). 
This series of experiments substantiates that the locality problem indeed stems from the characteristics of window attention.
%%%This group of experiments proves that the locality of window attention is the cause of the locality problem.

%\begin{figure}[htbp]

% \vspace{-0.5em}
\subsection{Influence of Window Size for Locality Problem}\label{section: Window Size}
% \vspace{-0.5em}
The window size in window attention, akin to the receptive field in CNN, delineates the computational scope. To examine the impact of window size, assuming the locality of window attention causes the locality problem, we investigate three scenarios: window sizes of 4, 7, and 14 for the last stage. The outcomes of these experiments are depicted in Fig.~\ref{fig3} (b). Notably, relatively small blocks emerge with a window size of 4 (column 4), while a shifted window's rough outline materializes with a window size of 14 (column 5). This series of experiments underscores the pivotal role of window size in the locality problem.
%%%The window size in window attention is similar to the receptive field in CNN, which represents the range that computation is concerned. Therefore, various window sizes should result in different stylized images, if the locality of window attention is the cause of the locality problem. We test the three cases when the window size of the last stage is 4, 7, or 14. Fig.~\ref{fig3} (b) shows the results of experiments. Relatively small blocks can be found when the window size of the last stage is 4 (column 4), while a rough shape of shifted window appears when the window size is set to 14 (column 5). This group of experiments proves that window size plays an important role in the locality problem.

% \vspace{-0.5em}
\subsection{Locality Phenomenon in Feature Maps}\label{section: Feature Maps}
\begin{figure}[t]
\centerline{\includegraphics[width=0.9\linewidth]{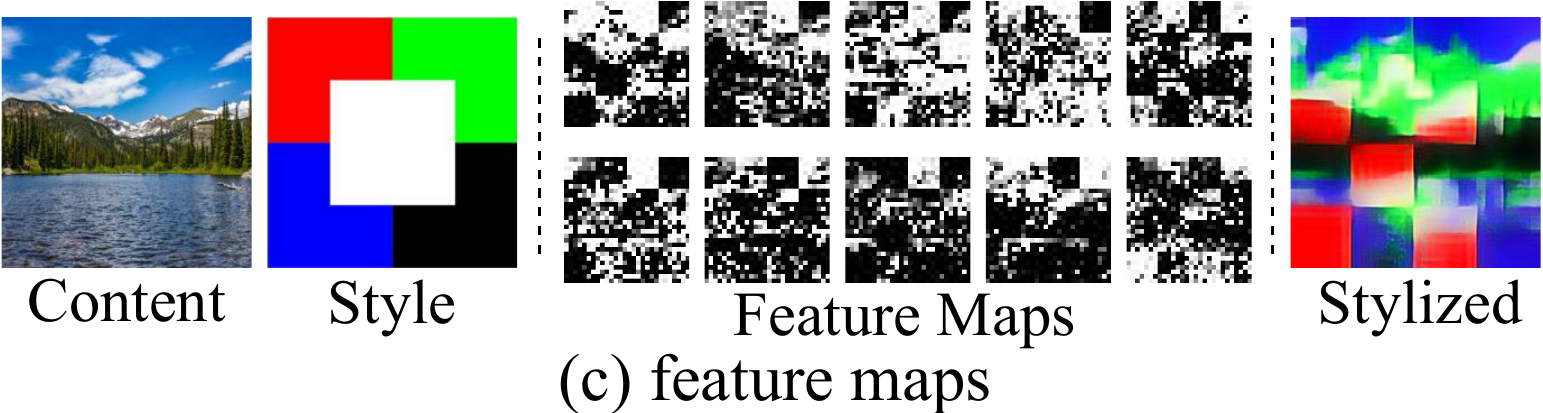}}
% \vspace{-0.1in}
\caption{The last feature maps from Swin-based encoder.}
\label{fig4}
% \vspace{-0.5em}
\end{figure}

% \vspace{-0.5em}
% In Section \ref{section: Location of Global MSA} and \ref{section: Window Size}, we discuss the changes in external factors, and we will give a shot at internal factors in Section \ref{section: Feature Maps} and \ref{section: Attention Maps}. Since the basis of the transformer-based transfer module is the similarity between content and style features, the content features should leave clues if the stylized images are grid-like. For this reason, we check out all the feature maps from the last layer of the encoder and list some of them (see Fig.~\ref{fig4}), which are convincing evidence to prove that features from window attention have strongly localized.
In previous parts, we discuss the changes in external factors, and we will give a shot at internal factors in the following parts. Since the basis of the transformer-based transfer module is the similarity between content and style features, the content features should leave clues if the stylized images are grid-like. For this reason, we check out all the feature maps from the last layer of the encoder and list some of them (see Fig.~\ref{fig4}), which are convincing evidence to prove that features from window attention have strongly localized.

%\begin{figure}[htbp]

%\begin{figure}[htbp]
\begin{figure}[t]
\centerline{\includegraphics[width=0.9\linewidth]{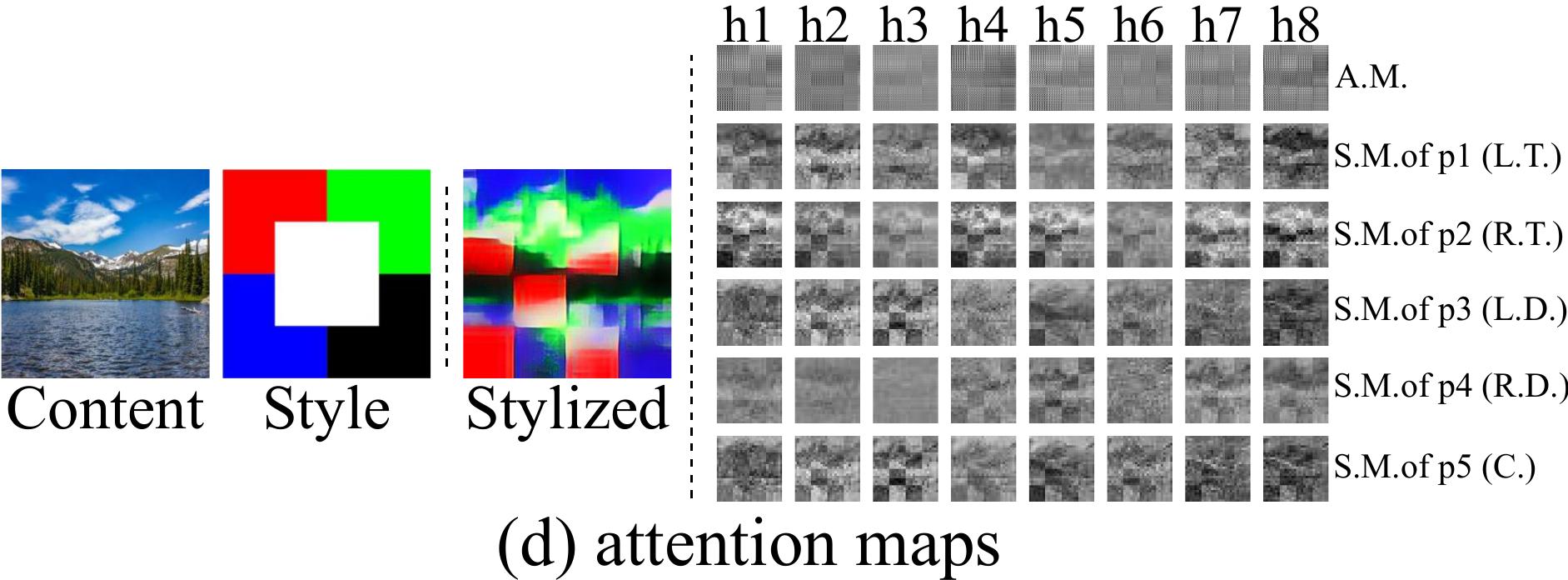}}
% \vspace{-0.1in}
\caption{Attention maps (row 1) and similarity maps (rows 2-6) of the five points. Attention and similarity maps differ in shape. They are scaled for easy observation. And $hi$ denotes $i$-th attention head. S/A.M. is short for ``similarity/attention maps" and L/R/T/D/C for ``left/right/top/down/center”.}
% \vspace{-1.5em}
\label{fig5}
\end{figure}

% \vspace{-0.5em}
\subsection{Locality Phenomenon in Attention Maps}\label{section: Attention Maps}
% \vspace{-0.5em}
To highlight the adverse impact of content feature locality on stylization, we analyze attention maps from the first inter-attention \cite{p79} in the transfer module (see Fig.~\ref{fig5}). Five points, representing corners (p1: top-left in red, p2: top-right in green, p3: bottom-left in blue, p4: bottom-right in black), and the central point (p5: white) are selected from style features to gauge their similarity with content features. These points, extracted from specific columns of attention maps and reshaped into squares, mirror content feature shapes. The similarity map of p1 reveals pronounced responses aligned with red blocks in the stylized image. Conversely, p2, p3, and p5 exhibit robust responses in areas devoid of red blocks.  As for p4's similarity map, responses are distributed widely. These outcomes underline the propagation of window attention's locality from content features within the encoder to the attention maps of the transfer module. This influence significantly disrupts the stylization process, ultimately culminating in the locality problem. To address this issue, we present the SpW Attention and S2WAT solutions.

\section{Method}
\begin{figure*}[htbp]
\centerline{\includegraphics[width=1.0\linewidth]{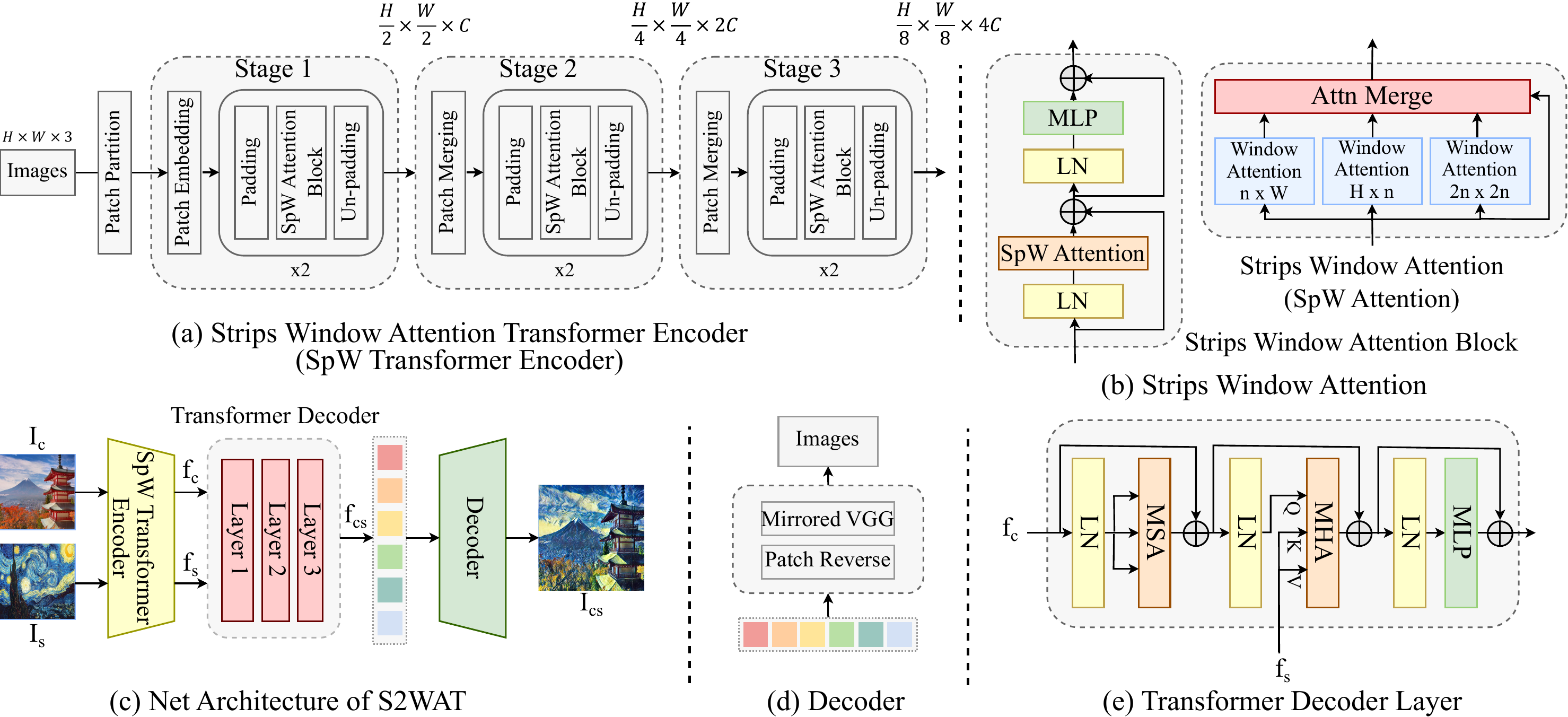}}
% \caption{Overall pipeline of the proposed S2WAT. (a) the structure of encoder, a hierarchical network with three scaling stages to extract features from inputs; (b) SpW Attention, where results of window attentions are adaptively merged by Attn Merge; (c) the overall pipeline of S2WAT; (d) the decoder of mirrored VGG; (e) the transformer layer used in the transfer module.}
\caption{Overall pipeline of the proposed S2WAT. Given a content image $I_c$ and a style image $I_s$, the encoder produces corresponding features $f_c$ and $f_s$. These features undergo style transfer from $f_s$ to $f_c$ within the transfer module, yielding stylized features $f_{cs}$. Subsequently, stylized features are decoded in the decoder to generate the stylized image $I_{cs}$.}
\label{fig6}
% \vspace{-0.1in}
\end{figure*}

Fig.~\ref{fig6} (c) presents the workflow of proposed S2WAT.
% As shown in Fig.~\ref{fig6} (c), the proposed S2WAT employs an encoder-transfer-decoder architecture.
% As shown in Fig.~\ref{fig6} (c), the proposed S2WAT employs an encoder-transfer-decoder architecture. Given a content image $I_c$ and a style image $I_s$, the encoder produces corresponding features $f_c$ and $f_s$. These features undergo style transfer from $f_s$ to $f_c$ within the transfer module, yielding stylized features $f_{cs}$. Subsequently, stylized features are decoded in the decoder to generate the stylized image $I_{cs}$.
%%%As depicted in Fig.~\ref{fig6} (c), the proposed S2WAT adopts the encoder-transfer-decoder architecture. Provided with a content image $I_c$ and a style image $I_s$, the corresponding features  $f_c$ and $f_s$ will be generated from encoder. After transferring the style characteristics on $f_s$ to $f_c$ in transfer module, we can obtain stylized features  $f_{cs}$ and finally receive the stylized image $I_{cs}$ by decoding $f_{cs}$ in decoder.

% In this part, we plan to discuss our solution, SpW Attention, first in Section \ref{section: Strips Window Attention}. And then, we introduce the overall architecture of the proposed S2WAT. After that, optimization strategy will be discussed in Section \ref{section: Network Optimization}.

% \vspace{-0.5em}
\subsection{Strips Window Attention}\label{section: Strips Window Attention}
% \vspace{-0.5em}
As illustrated in Fig.~\ref{fig6} (b), SpW Attention comprises two distinct phases: a window attention phase and a fusion phase.
%%As depicted in Fig.~\ref{fig6} (b), SpW Attention consists of two phases: a phase of window attention and a phase of fusion.

\begin{figure}[htbp]
\centerline{\includegraphics[width=1.0\linewidth]{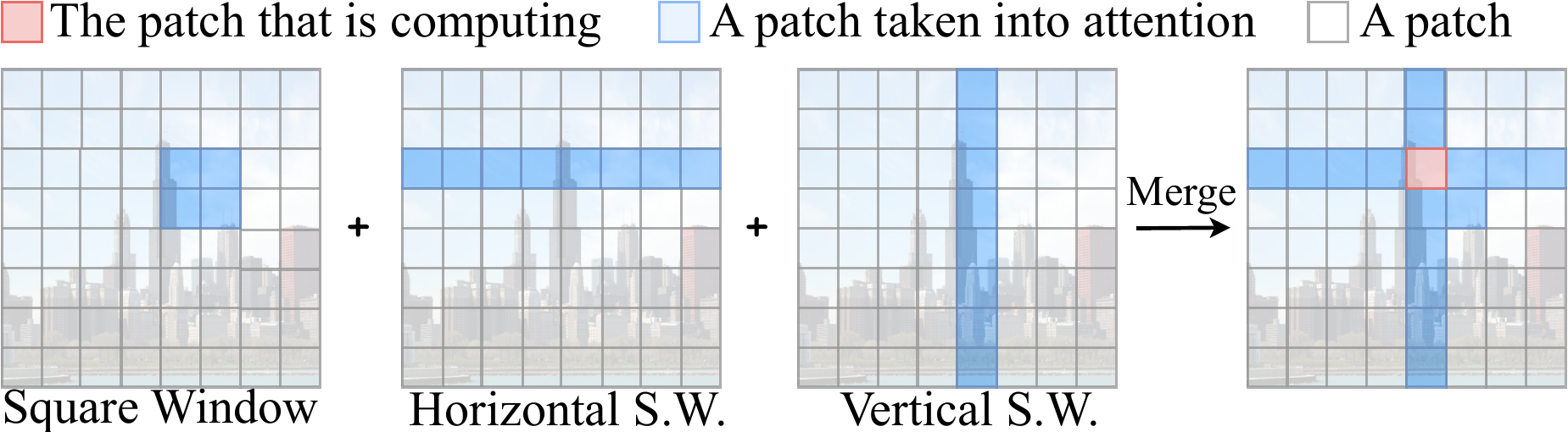}}
\caption{Receptive field of Strips Window Attention. Single strip-like window attention or square window attention can only glean information from limited targets in the image, while the merged one enlarges the receptive region to multiple directions. S.W. denotes ``strip window".}
\label{fig7}
% \vspace{-0.15in}
\end{figure}

% \subsubsection{Window Attention}
%%{\flushleft \bf Window Attention}.
% \vspace{1mm}
\noindent\textbf{Window Attention.}
Assuming input features possess a shape of $C \times H \times W$ and $n$ denotes the strip width, the first phase involves three distinct window attentions: a horizontal strip-like window attention with a window size of $n \times W$, a vertical strip-like window attention with a window size of $H \times n$, and a square window attention with a window size of $M \times M$ (where $M=2n$). A single strip-like window attention captures local information along one axis while accounting for long-range dependencies along the other. In contrast, the square window attention focuses on the surrounding information. Combining the outputs of these window attentions results in outputs that consider both local information and long-range dependencies. Illustrated in Fig.~\ref{fig7}, the merged approach gathers information from a broader range of targets, striking a favorable balance between computational complexity and the ability to sense global information.
%%%Supposing input features with shape of $ C \times H \times W $ and $ n $ is the strip width, the first phase is composed of three window attentions: a horizontal strip-like window attention with window size in shape of $ n \times W $, a vertical strip-like window attention with window size in shape of $ H \times n $, and a square one with window size in shape of $ M \times M \, (M=2n)$. Single strip-like window attention can sense local information on one axis while capturing long-range dependencies on another and square window attention focuses on the information surrounding by. Thus if we combine the results of these window attentions, the outputs will be the ones that consider not only the local information but also the long-range dependencies. As illustrated in Fig.~\ref{fig7}, the results of single strip-like window attention or square window attention can only obtain the information from limited targets in the image while the merged one receives the information from almost all kinds of targets. This is a beneficial trade-off between computational complexity and the ability to sense global information.

In computing square window attention, we follow \cite{p62} to include relative position bias $ B\in \mathbb{R}^{M^2 \times M^2} $ to each head in computing the attention map, as
%. The computing process is: 
\begin{equation}
\small
\textrm{W\mbox{-}MSA}_{M \times M}(Q,K,V) = \textrm{Softmax}(\frac{QK^T}{\sqrt{d}}+B)V, 
\label{eq1}
\end{equation}
where $ Q,K,V \in \mathbb{R}^{M^2 \times d} $ are the \textit{query}, \textit{key}, and \textit{value} matrices; $d$ is the dimension of \textit{query}/\textit{key}, $ M^2 $ is the number of patches in the window, and $ \textrm{W\mbox{-}MSA}_{M \times M} $ denotes multi-head self-attention using window in shape of $ M \times M $. 
We exclusively apply relative position bias to square window attention, as introducing it to strip-like window attention did not yield discernible enhancements.
%We only introduce relative position bias to square window attention and not to strip-like window attention for no apparent improvements appears after equipping it on. 

\begin{figure}[t]
\centerline{\includegraphics[width=1.0\linewidth]{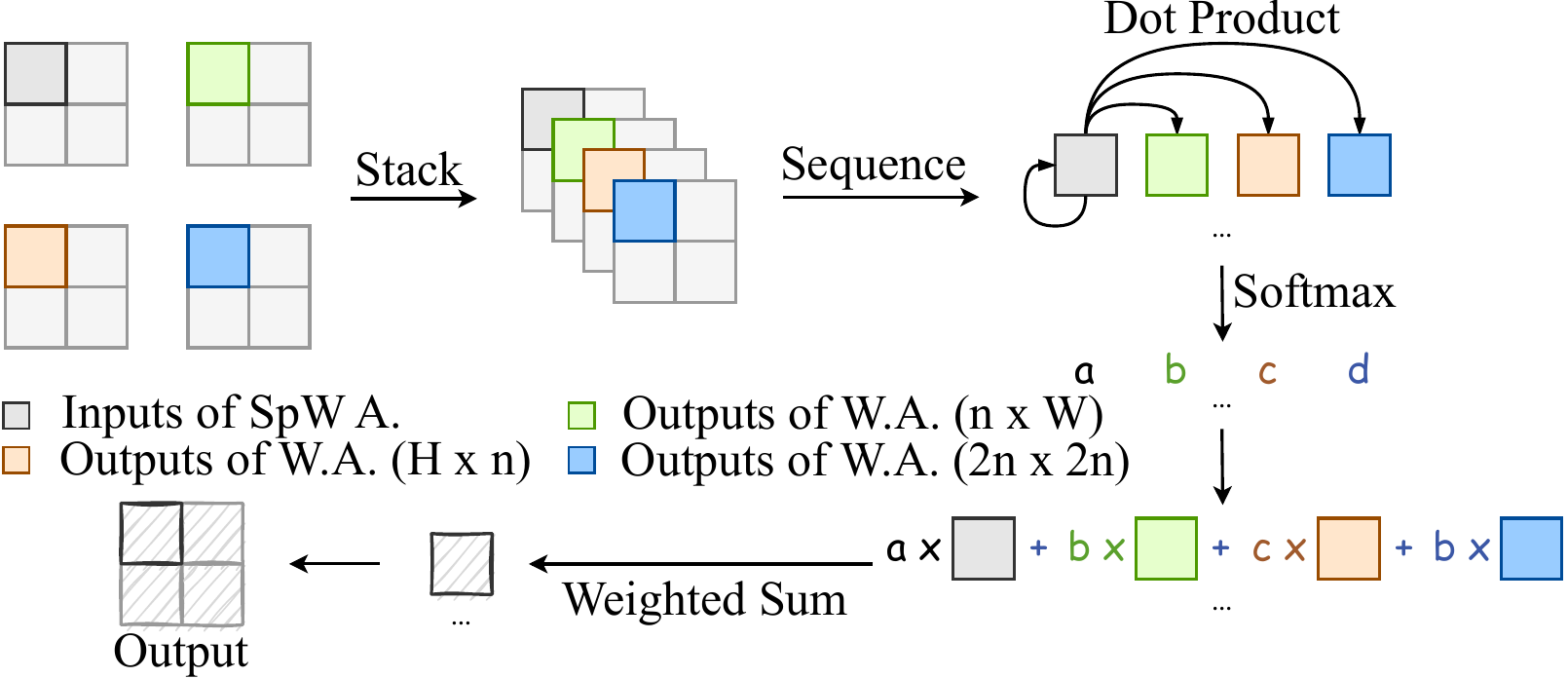}}
\caption{Workflow of ``Attn Merge". W./A. denotes ``Window/Attention".}
\label{fig8}
% \vspace{-0.1in}
\end{figure}

% \subsubsection{Attn Merge}
%%{\flushleft \bf Attn Merge}.
% \vspace{1mm}
\noindent\textbf{Attn Merge.}
%%After receiving the results from window attention phase, a fusion module named ``Attn Merge" is required to merge the results and inputs. As depicted in Fig.~\ref{fig8},  ``Attn Merge" is composed of three main steps. First, stacking the tensors; Second, computing similarity between the first tensor and the others at each spatial location; Third, computing weighted sum according to the similarity. ``Attn Merge" can also be computed efficiently as:
Following the completion of the window attention phase, a fusion module named ``Attn Merge" is engaged to consolidate the outcomes with the inputs. Illustrated in Fig.~\ref{fig8}, ``Attn Merge" comprises three core steps: first, tensor stacking; second, similarity computation between the first tensor and the rest at every spatial location; third, weighted summation based on similarity. The computational efficiency of ``Attn Merge" is noteworthy, as
\begin{equation}
    \begin{aligned}
        & Y = \textrm{Stack}(x, a, b, c), \quad Y \in \mathbb{R}^{n \times 4 \times d}, \\
	& x' = \textrm{Unsqueeze}(x), \quad x \in \mathbb{R}^{n \times 1 \times d}, \\
	& Z = x'Y^{T}Y, \quad Z \in \mathbb{R}^{n \times 1 \times d},  \\
	& z = \textrm{Squeeze}(Z), \quad z \in \mathbb{R}^{n \times d}, 
    \end{aligned}
    \label{eq2}
\end{equation}
%%\begin{align}
%%	& Y = \textrm{Stack}(x, a, b, c), \quad Y \in \mathbb{R}^{n \times 4 \times d}, \notag \\
%%	& x' = \textrm{Unsqueeze}(x), \quad x \in \mathbb{R}^{n \times 1 \times d}, \notag \\
%%	& Z = x'Y^{T}Y, \quad Z \in \mathbb{R}^{n \times 1 \times d}, \label{eq2} \\
%%	& z = \textrm{Squeeze}(Z), \quad z \in \mathbb{R}^{n \times d}, \notag
%%\end{align}
where $ x,a,b,c \in \mathbb{R}^{n \times d} $ are input tensors and $z$ is the outputs; $\textrm{Stack}$ denotes the operation to collect tensors in a new dimension and $\textrm{Unsqueeze} \,/\, \textrm{Squeeze} $ represents the operation to add or subtract a dimension of tensor. 
%%%%%``Attn Merge" is simple but effective in fusing tensors. As shown in ablation study (Section \ref{section: Ablation Study}), ``Attn Merge" is better than traditional sum and concatenating operations.
%%traditional sum and concatenating operations can not work well while ``Attn Merge" still works normally.

% \subsubsection{Strips Window Attention Block}
%{\flushleft \bf Strips Window Attention Block}.
% \vspace{1mm}
\noindent\textbf{Strips Window Attention Block.}
We now provide an overview of the comprehensive workflow of the SpW Attention block. The structure of the SpW Attention block mirrors that of a standard transformer block, except for the substitution of MSA with a SpW Attention (SpW-MSA) module. As depicted in Fig.~\ref{fig6} (b), a SpW Attention block comprises a SpW-MSA module, succeeded by a two-layer MLP featuring GLUE as the non-linear activation in between. Preceding each SpW-MSA module and MLP, a LayerNorm (LN) operation is applied, and a residual connection is integrated after each module. The computation process of a SpW Attention block unfolds as follows:
%%Here we take a look at the overall workflow of SpW Attention block. SpW Attention block has the same structure as the standard transformer block except for replacing the MSA with a SpW Attention (SpW-MSA) module. As illustrated in Fig.~\ref{fig6} (b), a SpW Attention block is composed of a SpW-MSA module, followed by a two-layer MLP with GLUE as non-linear activation in between. Before each SpW-MSA module and MLP, a LayerNorm (LN) is applied and a residual connection is attached after each module. The calculation process of SpW Attention block is as follows:
%\begin{gather}
\begin{equation}
    \begin{aligned}
    \small
	&\hat{z}^{l}_{n \times W} = \textrm{W\mbox{-}MSA}_{n \times W}(\textrm{LN}(z^{l-1})),  \\
	&\hat{z}^{l}_{H \times n} = \textrm{W\mbox{-}MSA}_{H \times n}(\textrm{LN}(z^{l-1})), \\
	&\hat{z}^{l}_{2n \times 2n} = \textrm{W\mbox{-}MSA}_{2n \times 2n}(\textrm{LN}(z^{l-1})),  \\
	%\tilde{z}^{l} = AttnMerge(LN(z^{l-1}),\, \hat{z}^{l}_{n \times W},\, \hat{z}^{l}_{H \times n},\, \hat{z}^{l}_{2n \times 2n}) + z^{l-1}  \notag \\
 &\tilde{z}^{l} = \mathcal{A}(\textrm{LN}(z^{l-1}),\, \hat{z}^{l}_{n \times W},\, \hat{z}^{l}_{H \times n},\, \hat{z}^{l}_{2n \times 2n}) + z^{l-1},   \\
	&z^{l} = \textrm{MLP}(\textrm{LN}(\hat{z}^{l}))+\tilde{z}^{l},\\
    \end{aligned}
    \label{eq3}
\end{equation}
%\end{gather}
where ``$\mathcal{A}$" means ``Attn Merge", $ z^{l} $ , $ \tilde{z}^{l} $ and $ \hat{z}^{l} $ denote the outputs of MLP, ``Attn Merge", and W-MSA for block $l$, respectively; $ \textrm{W\mbox{-}MSA}_{n \times m} $ denotes multi-head self-attention using window in shape of $ n \times m $. As shown in \eqref{eq3}, the SpW Attention block primarily consists of two parts: SpW Attention (comprising W-MSA and ``Attn Merge") and an MLP.
%%the SpW Attention block is mainly composed of two parts: SpW Attention (consisting of W-MSA and ``Attn Merge") and MLP.

% \subsubsection{Computational Complexity}
%{\flushleft \bf Computational Complexity}.
% \vspace{1mm}
\noindent\textbf{Computational Complexity.}
To make the cost of computation in SpW Attention clear, we compare the computational complexity of MSA, W-MSA, and the proposed SpW-MSA. Supposing the window size of W-MSA and the strip width of SpW-MSA are equal to $ M $ and $ C $ is the dimension of inputs, the computational complexity of a global MSA, a square window based one, and a Strips Window based one on an image of $ h \times w $ patches are:
\begin{align}
	\Omega(\textrm{MSA})              =\, & 2(wh)^{2}C + 4whC^{2}, \label{eq4} \\
	\Omega(\textrm{W \mbox{-} MSA})   =\, & 2M^{2}whC + 4whC^{2}, \label{eq5} \\
	\Omega(\textrm{SpW \mbox{-} MSA}) =\, & 2M(w^{2}h+wh^{2}+4Mwh)C + \notag \\
                                 & 12whC^{2} + 8whC. \label{eq6}    
\end{align}
As shown in Eqs. \eqref{eq4}-\eqref{eq6}, MSA is quadratic to the patch number $ hw $, and W-MSA  is linear when $ M $ is fixed. And the proposed SpW-MSA is something in the middle.

\subsection{Overall Architecture}
%%%Different from StyTr2 \cite{p45} which has two individual encoders for encoding inputs from different domains, we follow the traditional encoder-transfer-decoder architecture of UST to encode content and style images in one encoder. An overview of architecture is presented in Fig.~\ref{fig6}.
In contrast to StyTr2~\cite{p45}, which employs separate encoders for different input domains, we adhere to the conventional encoder-transfer-decoder design of UST. This architecture encodes content and style images using a single encoder. An overview is depicted in Fig.~\ref{fig6}.

% \subsubsection{Encoder}
%%{\flushleft \bf Encoder}.
% \vspace{1mm}
\noindent\textbf{Encoder.}
Like Swin, S2WAT's encoder initially divides content and style images into non-overlapping patches using a patch partition module. These patches serve as ``tokens" in transformers. We configure the patch size as $2 \times 2$, resulting in patch dimensions of $2 \times 2 \times 3 = 12$. Subsequently, a linear embedding layer transforms the patches into a user-defined dimension ($C$).
%%Similar to Swin, the encoder of S2WAT first splits the content and style images into non-overlapping patches by a patch partition module. These patches were treated as the ``tokens" in transformer \cite{p48} or the features in CNNs. We set the patch size to $2 \times 2$ and thus the dimension of patches is $ 2 \times 2 \times 3 = 12 $. Then, a linear embedding layer is applied on the patches to project them to an arbitrary dimension (denote as $ C $).

%%After the patch embedding layer, 
After embedding, the patches proceed through a series of consecutive SpW Attention blocks, nestled between padding and un-padding operations. Patches are padded to achieve divisibility by twice the strip width and cropped (un-padded) after SpW Attention blocks, preserving the patch count. Notably, patch padding employs reflection to mitigate potential light-edge artifacts that can arise when using constant 0 padding. These SpW Attention blocks uphold the patch count ($\frac{H}{2} \times \frac{W}{2}$) and, in conjunction with the patch embedding layer and padding/un-padding operations, constitute ``Stage 1".
%%%After embedding, the patches will be sent to several successive SpW Attention blocks which are sandwiched between a pair of padding \& un-padding operations. Patches will be padded to be divisible by twice the strip width and be cropped (un-padded) after SpW Attention blocks to keep the number of patches unchanged. One thing that needs attention is that we pad the patches by reflecting for light-edge results may appear when padding with constant 0. The SpW Attention blocks maintain the number of patches ($ \frac{H}{2} \times \frac{W}{2} $), and together with the patch embedding layer and padding \& un-padding operations are referred to as ``Stage 1".

To achieve multi-scale features, gradual reduction of the patch count is necessary as the network deepens. Swin introduces a patch merging layer as a down-sample module, extracting elements with a two-step interval along the horizontal and vertical axes. By concatenating $2 \times 2$ groups of these features in the channel dimension and reducing channels from $4C$ to $2C$ through linear projection, a 2x downsampling result is obtained. Subsequent application of SpW Attention blocks, flanked by padding and un-padding operations, transforms the features while preserving a resolution of $\frac{H}{4} \times \frac{W}{4}$. This combined process is designated as ``Stage 2". This sequence is reiterated for ``Stage 3", yielding an output resolution of $\frac{H}{8} \times \frac{W}{8}$. Consequently, the encoder's hierarchical features in S2WAT can readily be employed with techniques like FPN or U-Net.
%%%To produce multi-scale features, the number of patches needs to be reduced progressively as the network gets deeper. The patch merging layer in Swin is introduced as a down-sample module, which extracts elements from features with an interval of two on both horizontal and vertical axes. After concatenating the $ 2 \times 2 $ groups of extracted features on the dimension of channels and reducing the $ 4C $ channels to $ 2C $ by a linear projection, we obtain the results of 2x downsampling. Sandwiched between padding \& un-padding, SpW Attention blocks are applied afterwards to transform features, which keep the resolution at $ \frac{H}{4} \times \frac{W}{4} $. The patch merging layer together with SpW Attention blocks and padding \& un-padding are denoted as ``Stage 2". This procedure repeats once more as ``Stage 3" with output resolutions of $ \frac{H}{8} \times \frac{W}{8} $. As a result, hierarchical features from the encoder of S2WAT can be easily used when applying techniques like FPN or U-Net.

% \subsubsection{Transfer Module}
%%{\flushleft \bf Transfer Module}.
% \vspace{1mm}
\noindent\textbf{Transfer Module.}
A multi-layer transformer decoder replaces the transfer module, similar to StyTr2~\cite{p45}. In our implementation, we maintain a close resemblance to the original transformer decoder~\cite{p48}, with two key distinctions from StyTr2: a) The initial attention module of each transformer decoder layer is MSA, whereas StyTr2 employs MHA (multi-head attention); b) LayerNorm precedes the attention module and MLP, rather than following them. The structure is presented in Fig.~\ref{fig6} (e) and more details can be found in codes.
\noindent\textbf{Decoder.}
% In line with prior research~\cite{p21,p37,p41}, we utilize a mirrored VGG for decoding stylized features. However, due to the $ \frac{WH}{64} \times 4C $ sequence shape, a reshaping step is necessary for the outputs of the transfer module. As illustrated in Fig.~\ref{fig6} (d), preceding the mirrored VGG is a patch reverse layer for this reshaping process. Subsequent to the mirrored VGG, we attain the ultimate results in a $3 \times H \times W$ resolution. Additional details can be found in the Appendix.
In line with prior research~\cite{p21,p37,p41}, we utilize a mirrored VGG for decoding stylized features. Detailed implements are available in codes.
%%%We follow previous works \cite{p21,p37,p41} to leverage a mirrored VGG to decode the stylized features. However, because of the shape of $  \frac{WH}{64} \times 4C $ sequences, the outputs of transfer module need to be reshaped first. As depicted in Fig.~\ref{fig6} (d), before the mirrored VGG, there is a patch reverse layer to reshape the outputs of transfer module. After the mirrored VGG, we obtain the final results in a resolution of $ 3 \times H \times W $. Details can be found in Appendix.

% \vspace{-0.5em}
\subsection{Network Optimization}\label{section: Network Optimization}
% \vspace{-0.5em}
Similar to \cite{p21}, we formulate two distinct perceptual losses for gauging the content dissimilarity between stylized images $I_{cs}$ and content images $I_{c}$, along with the style dissimilarity between stylized images $I_{cs}$ and style images $I_{s}$.
The content perceptual loss is defined as:
%%Similar to \cite{p21}, we construct two different perceptual losses to measure the content difference between stylized images $ I_{cs} $ and content images $ I_{c} $, as well as the style difference between stylized images $ I_{cs} $ and style images $ I_{s} $. The content perceptual loss is defined as:
\begin{gather}
\mathcal{L}_{content} = \sum_{l \in \mathcal{C}} \Vert \overline{\phi_{l}(I_{cs})} - \overline{\phi_{l}(I_{c})} \Vert_{2}^{}, \label{eq8}
\end{gather}
where the overline denotes mean-variance channel-wise normalization; $ \phi_{l}(\cdot) $ represents extracting features of layer $ l $ from a pre-trained VGG19 model; $ \mathcal{C} $ is a set consisting of $ relu4\_1 $ and $ relu5\_1 $ in the VGG19. The style perceptual loss is defined as:
\begin{equation}
\begin{aligned}
\mathcal{L}_{style} = & \sum_{l \in \mathcal{S}} \Vert \mu(\phi_{l}(I_{cs})) - \mu(\phi_{l}(I_{s})) \Vert_{2} \\
& + \Vert \sigma(\phi_{l}(I_{cs})) - \sigma(\phi_{l}(I_{s})) \Vert_{2},
\end{aligned}
\label{eq9}
\end{equation}
where $ \mu(\cdot) $ and $ \sigma(\cdot) $ denote mean and variance of features, respectively; and $ \mathcal{S} $ is a set consisting of $ relu2\_1 $, $ relu3\_1 $, $ relu4\_1 $ and $ relu5\_1 $ in the VGG19.

We also adopt identity losses \cite{p37} to further maintain the structure of the content image and the style characteristics of the style image. The two different identity losses are defined as:
\begin{gather}
\mathcal{L}_{id1} = \Vert I_{cc} - I_{c}\Vert_{2}^{} + \Vert I_{ss} - I_{s}\Vert_{2}^{}, \label{eq10} \\
\resizebox{0.85\linewidth}{!}{
$\mathcal{L}_{id2} = \sum_{l \in \mathcal{N}} \Vert \phi_{l}(I_{cc}) - \phi_{l}(I_{c})\Vert_{2}^{} + \Vert \phi_{l}(I_{ss}) - \phi_{l}(I_{s})\Vert_{2}^{}$}, \label{eq11}
\end{gather}
where $ I_{cc} $ (or $ I_{ss} $) denotes the output image stylized from two same content (or style) images and $ \mathcal{N} $ is a set consisting of $ relu2\_1 $, $ relu3\_1 $, $ relu4\_1 $ and $ relu5\_1 $ in the VGG19. Finally, our network is trained by minimizing the loss function defined as:
\begin{equation}
    \resizebox{0.85\linewidth}{!}{
$\mathcal{L}_{total} = \lambda_{c}\mathcal{L}_{content} + \lambda_{s} \mathcal{L}_{style} + \lambda_{id1}\mathcal{L}_{id1} + \lambda_{id2} \mathcal{L}_{id2}$
},
\label{eq12}
\end{equation}
where $ \lambda_{c} $, $ \lambda_{s} $, $ \lambda_{id1} $, and $ \lambda_{id2} $ are the weights of different losses. We set the weights to 2, 3, 50, and 1, relieving the impact of  magnitude differences.

\begin{figure*}[htbp]
\centerline{\includegraphics[width=1.0\linewidth]{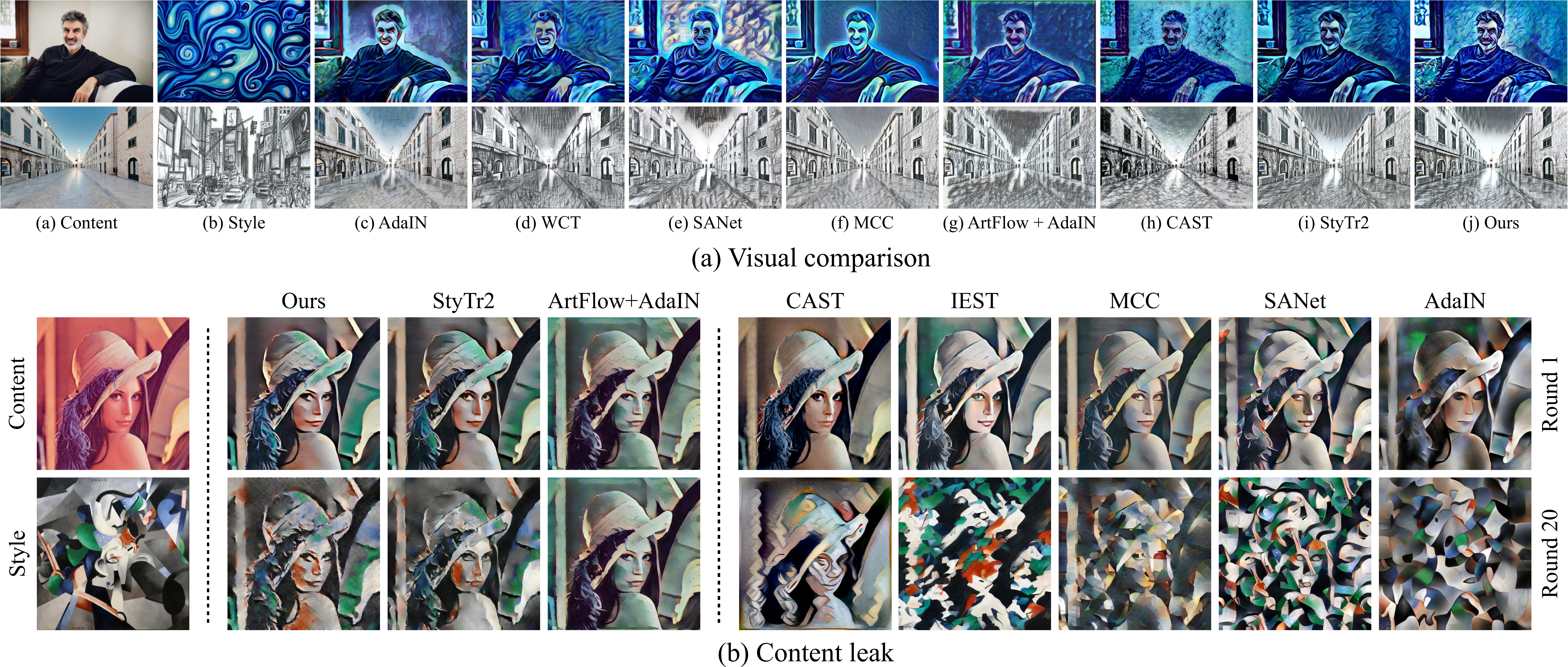}}
\caption{Visual comparison of the results from SOTA methods and visualization of content leak. A.F. denotes ``ArtFlow".} 
% \vspace{-0.1in}
\label{fig9}
\end{figure*}

\section{Experiments}
MS-COCO \cite{p84} and WikiArt \cite{p85} are used as the content dataset and the style dataset respectively. 
Other implementation details are available in Appendix and codes.
% \subsection{Implementation Details}
% % \subsubsection{Dataset}
% % {\flushleft \bf Dataset}.
% \noindent\textbf{Dataset.} We use MS-COCO \cite{p84} and WikiArt \cite{p85} as the content dataset and the style dataset respectively. In the stage of training, the images are resized to 512 on the shorter side first and then randomly cropped to $ 224 \times 224 $. In the stage of testing, images of any size are accepted.

% % \subsubsection{Training Information}
% % {\flushleft \bf Training Information}.
% \vspace{1mm}
% \noindent\textbf{Training Information.} We use Pytorch \cite{p86} framework to implement S2WAT and train it for 40000 iterations with a batch size of 4. An Adam optimizer \cite{p87} and the warmup learning rate adjustment strategy \cite{p88} are applied with the initial learning rate of 1e-4. Our model is trained on a single Tesla V100 GPU for approximately 10 hours.

\subsection{Style Transfer Results}\label{section: Style Transfer Results}
% \vspace{-0.5em}
% This section contrasts S2WAT with previous SOTAs, including AdaIN \cite{p21}, WCT \cite{p22}, SANet \cite{p37}, MCC \cite{p41}, ArtFlow \cite{p43}, IEST \cite{p42}, CAST \cite{p46}, StyTr2 \cite{p45} and InST \cite{p100}. 
In this Section, we compare the results between the proposed S2WAT and previous SOTAs, including AdaIN \cite{p21}, WCT \cite{p22}, SANet \cite{p37}, MCC \cite{p41}, ArtFlow \cite{p43}, IEST \cite{p42}, CAST \cite{p46}, StyTr2 \cite{p45} and InST \cite{p100}.

\noindent\textbf{Qualitative Comparison.}
In Fig.~\ref{fig9} (a), we present visual outcomes of the compared algorithms. AdaIN, relying on mean and variance alignment, fails to capture intricate style patterns. While WCT achieves multi-level stylization, it compromises content details. SANet, leveraging attention mechanisms, enhances style capture but may sacrifice content details. MCC, lacking non-linear operations, faces overflow issues. Flow-based ArtFlow produces content-unbiased outcomes but may exhibit undesired patterns at borders. CAST retains content structure through contrastive methods but may compromise style. InST's diffusion models yield creative results but occasionally sacrifice consistency. StyTr2 and proposed S2WAT strike a superior balance, with S2WAT excelling in preserving content details (e.g., numbers on the train, the woman's glossy lips, and letters on billboards), as highlighted in dashed boxes in Fig.~\ref{fig9} (a). Additional results are available in the Appendix.
\noindent\textbf{Quantitative Comparison.}
In this section, we follow a methodology akin to \cite{p21,p43,p45} utilizing losses as indirect metrics. Style, content, and identity losses serve as metrics, evaluating style quality, content quality, and input information retention, respectively. Additionally, inspired by \cite{p43}, the Structural Similarity Index (SSIM) is included to gauge structure preservation. As shown in Table \ref{table:1}, S2WAT achieves the lowest content and identity losses, while SANet exhibits the lowest style loss. StyTr2 and S2WAT show comparable loss performance, emphasizing style and content, respectively. Due to its content-unbiased nature, ArtFlow registers identity losses of 0, signaling an unbiased approach. While ArtFlow is unbiased, S2WAT outperforms it in style and SSIM. S2WAT attains the highest SSIM, indicating superior content structure retention. It excels in preserving both content input structures and artistic style characteristics simultaneously.

\begin{table*}[htbp]
    \begin{center}
         \resizebox{1.0\linewidth}{!}{
         \begin{tabular}{c c c c c c c c c c c c}
            \hline
            Method & Ours & InST & StyTr2 & CAST & IEST & ArtFlow\mbox{-}AdaIN & ArtFlow\mbox{-}WCT & MCC & SANet & WCT & AdaIN \\
            \hline
            \textit{Content Loss} $\downarrow$ & \pmb{1.66} & 3.73 & 1.83 & 2.07 & 1.81 & 1.93 & 1.73 & 1.92 & 2.16 & 2.56  & \underline{1.71}  \\
            \textit{Style Loss} $\downarrow$ & 1.74 & 29.98 & \underline{1.52} & 4.33 & 2.72 & 1.90 & 1.89 & 1.70 & \pmb{1.11} & 2.23  & 3.50 \\
            \textit{Identity Loss 1} $\downarrow$  & \pmb{0.16} & 0.71& \underline{0.26} & 1.94 & 0.91 & 0.00 & 0.00 & 1.07 & 0.81 & 3.01  & 2.54 \\
            \textit{Identity Loss 2} $\downarrow$ & \pmb{1.38} & 134.23 & \underline{3.10} & 18.72 & 7.16 & 0.00 & 0.00 & 7.72 & 6.03 & 21.88 & 17.97 \\
            \textit{SSIM} $\uparrow$ & \pmb{0.651} & 0.401 & 0.605 & \underline{0.619} & 0.551 & 0.578 & 0.612 & 0.578 & 0.448 & 0.364 & 0.539 \\
            % \textit{Time(seconds)} $\downarrow$ & 0.558 & 0.237 &  \pmb{0.042} & 0.061 & 0.325 & 0.329 & 0.078 & 0.061 & 0.590 & \pmb{0.042} \\
            \hline
         \end{tabular}
         }
         % \vspace{-4mm}
    \end{center}
    % \vspace{-0.1in}
    \caption{Quantitative evaluation results of different style transfer methods. The losses above are average values from 400 random samples, while SSIMs are computed average from 100 pieces. For a fair comparison, we take $ relu1\_1 $ into consideration in computing style loss and identity loss 2 while not in the training of S2WAT. The optimal results are highlighted in bold, the second-best results are underlined, and instances with a value of 0 are derived from unbiased methods.}
    \label{table:1}
    % \vspace{-0.2in}
\end{table*}

% \vspace{-0.5em}
\subsection{Content Leak}\label{section: Content Leak}
Content leak problem occur when applying the same style image to a content image repeatedly, especially if the model struggles to preserve content details impeccably.
Following \cite{p43,p45}, We investigate content leakage in the stylization process, focusing on S2WAT and comparing it to ArtFlow, StyTr2, CNN-based, and SDM-based methods. Our experiments, detailed in Fig.~\ref{fig9} (b), reveal S2WAT and StyTr2, both transformer-based, exhibit minimal content detail loss over 20 iterations, surpassing CNN and SDM methods known for noticeable blurriness. While CAST alleviates content leak partially, the stylized effect remains suboptimal. In summary, S2WAT effectively mitigates the content leak issue. 

InST occasionally underperforms, especially when content and style input styles differ significantly, potentially due to overfitting in the Textual Inversion module during single-image training. More details are available in the Appendix.

\subsection{Ablation Study}\label{section: Ablation Study}
\begin{figure}[t]
\centerline{\includegraphics[width=1.0\linewidth]{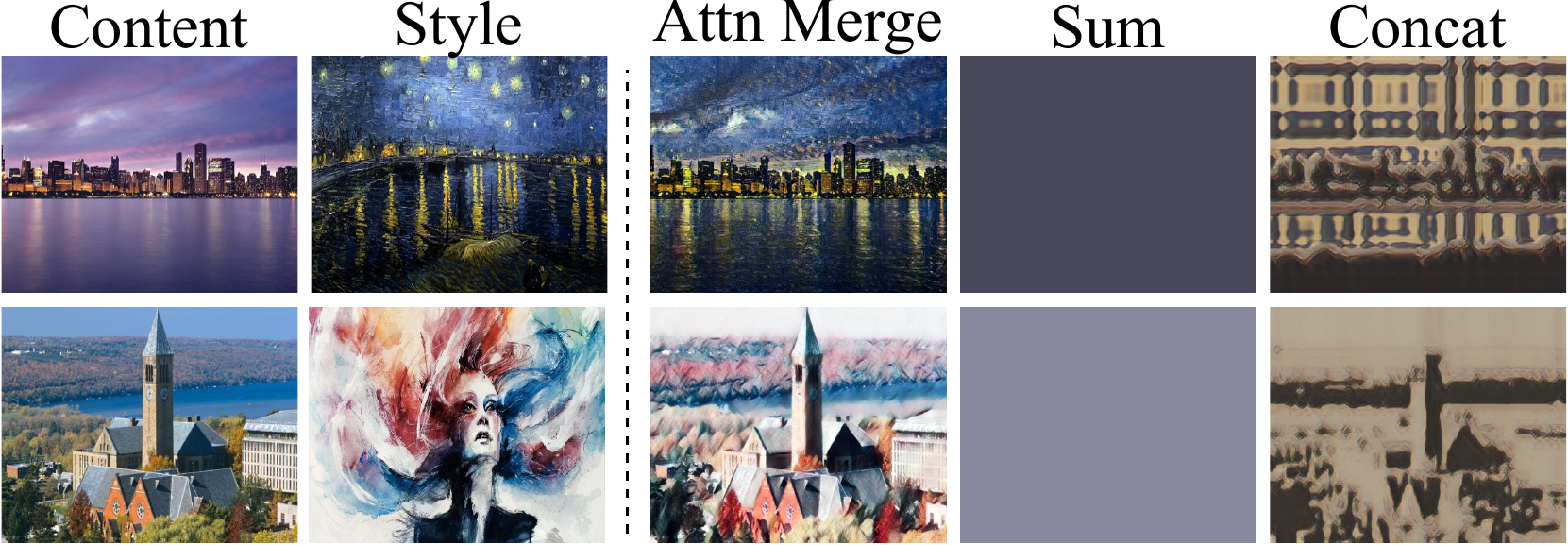}}
% \vspace{-0.1in}
\caption{Visual comparisons when utilizing different fusion strategies for attention outputs from multiple windows.}
\label{fig11}
% \vspace{-0.2in}
\end{figure}

% \subsubsection{Attn Merge}
% \noindent\textbf{Attn Merge.} 
%%{\flushleft \bf Attn Merge}.
\noindent\textbf{Attn Merge.}
In order to showcase the effectiveness and superiority of ``Attn Merge", we undertake experiments where ``Attn Merge" is replaced by fusion strategies such as the concatenation operation (as employed by CSWin) or the sum operation. The outcomes are depicted in Fig.~\ref{fig11}. Stylized images generated using the sum operation are extensively corrupted, indicating a failure in model optimization. On the other hand, outputs obtained through concatenation relinquish a substantial portion of information from input images, particularly the style images.
An intuitive rationale for this phenomenon lies in the optimization challenges posed by straightforward fusion operations. Comprehensive explanations are available in the Appendix. The proposed ``Attn Merge", however, facilitates smooth information transmission, allowing the model to undergo normal training.

%%%%To demonstrate the effectiveness and superiority of ``Attn Merge", we conduct experiments that replace ``Attn Merge" with fusion strategies like the concatenating operation (taken by CSWin) or the sum operation. Fig.~\ref{fig11} presents the results. Stylized images based on the sum operation are totally corrupted, which means the model fails to be trained in optimization, while the outputs from concatenating lose the majority of information from input images, especially the style images. 
%However, 
%%%%One intuitive explanation for the phenomenon is the optimization obstacles due to simple fusion operations. Detailed descriptions can be found in Appendix. Under the proposed ``Attn Merge", the model can be trained normally with information transmitting smoothly.
% and the information from input images are able to be transmitted smoothly. 

\begin{figure}[t]
\centerline{\includegraphics[width=1.0\linewidth]{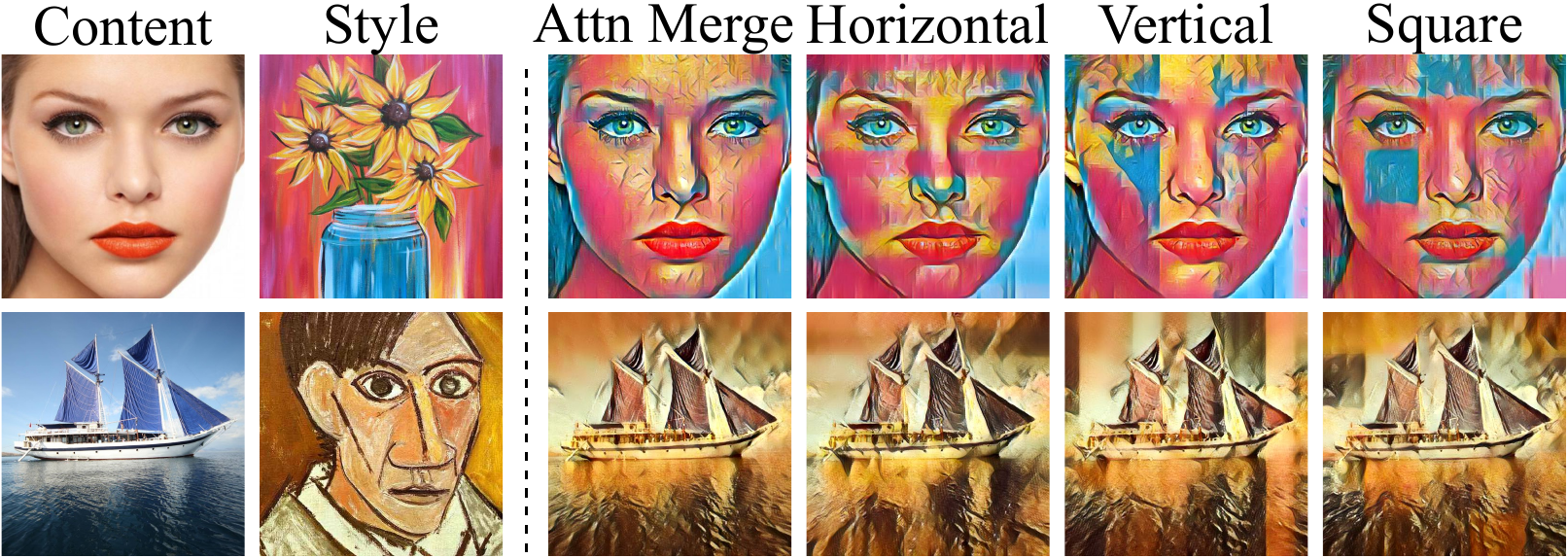}}
\caption{Visual comparisons for the ablation study when employing different window attention mechanisms.}
\label{fig12}
% \vspace{-0.5em}
\end{figure}

% \subsubsection{Strips Window}
% \noindent\textbf{Strips Window.} 
%%{\flushleft \bf Strips Window}.
% \vspace{1mm}
\noindent\textbf{Strips Window.}
To verify the demand to fuse outputs from window attention of various sizes, we carry out experiments employing window attention with distinct window sizes independently. As illustrated in Fig.~\ref{fig12}, the utilization of horizontal or vertical strip-like windows in isolation yields corresponding patterns. Applying square windows alone results in grid-like patterns. However, the incorporation of ``Attn Merge" to fuse outcomes leads to pleasing stylized images, surpassing the results obtained solely from window attention.
Further details regarding the ablation study for Swin and Swin with Mix-FFN can be found in the Appendix.
%%%To prove the necessity of fusing the results from window attention of different window sizes, we conduct experiments applying the window attention with different window sizes individually. As depicted in Fig.~\ref{fig12}, horizontal or vertical strip-like patterns appear when applying the window attention with horizontal or vertical strip-like windows separately. For square windows, grid-like patterns are produced. Nevertheless, stylized images will be delightful if leveraging ``Attn Merge" to fuse the results rather than applying the window attention alone.
%%%Ablation study for Swin and Swin with mix-FFN can be found in Appendix.

% \vspace{-0.5em}
\subsection{User Study}
In comparing virtual stylization effects between S2WAT and the aforementioned SOTAs like StyTr2, ArtFlow, MCC, and SANet, user studies were conducted. Using a widely-employed online questionnaire platform, we created a dataset comprising 528 stylized images from 24 content images and 22 style images. Participants, briefed on image style transfer and provided with evaluation criteria, assessed 31 randomly selected content and style combinations. Criteria emphasized preserving content details and embodying artistic attributes. With 3002 valid votes from 72 participants representing diverse backgrounds, including high school students and professionals in computer science, art, and photography, our method achieved a marginal victory in the user study, as reflected in Table~\ref{table:2}. Additional details including an example questionnaire page can be found in the Appendix.

\begin{table}[t]
    \begin{center}
         \resizebox{1.0\linewidth}{!}{
         \begin{tabular}{c c c c c c}
            \hline
            Method & Ours & StyTr2 & ArtFlow & MCC & SANet \\
            \hline
            \textit{Percent}(\%)  & \pmb{25.4} & 23.6 & 13.3 & 19.4 & 18.3 \\
            \hline
         \end{tabular}
         }
    \end{center}
    % \vspace{-4mm}
    % \vspace{-0.1in}
    \caption{Percentage of votes in the user study.}
    % \vspace{-0.2in}
    % \vspace{-1.7em}
    \label{table:2}
\end{table}

% \vspace{-0.3em}
\section{Conclusion}
In this paper, we introduce S2WAT, a pioneering image style transfer framework founded upon a hierarchical vision transformer architecture. 
S2WAT's prowess lies in its capacity to simultaneously capture local and global information through SpW Attention. The SpW Attention mechanism, featuring diverse window attention shapes, ensures an optimal equilibrium between short- and long-range dependency modeling, further enhanced by our proprietary ``Attn Merge". This adaptive merging technique efficiently gauges the significance of various window attentions based on target similarity.
Furthermore, S2WAT mitigates the content leak predicament, yielding stylized images endowed with vibrant style attributes and intricate content intricacies.

%%%In this paper, we propose S2WAT, a novel image style transfer framework based a on hierarchical vision transformer. S2WAT consists of three parts: a) an encoder based on a three-stage hierarchical vision transformer, b) a transfer module based on a multi-layer transformer decoder, c) and a decoder of mirrored VGG. S2WAT is able to capture both local and global information by SpW Attention. In SpW Attention, various shapes of window attentions allow the optimal balance between short- and long-range dependencies modeling, and they can be merged by our designed ``Attn Merge". ``Attn Merge" can adaptively decide the importance of different window attentions according to the similarity of the target. An issue called locality problem is revealed and we address it by SpW Attention. Moreover, S2WAT can alleviate the content leak problem and produce stylized images with vivid style characteristics and fine content details.

\section{Acknowledgement}
This work was supported by the National Key R\&D Program of China (2021YFB3100700), the National Natural Science Foundation of China (62076125, U20B2049, U22B2029, 62272228), and Shenzhen Science and Technology Program (JCYJ20210324134408023, JCYJ20210324134810028).

%%%%%%%%% REFERENCES
% {\small
% % we modified ieee_fullname.bst to avoid name sorting rule.
% \bibliographystyle{ieee_fullname}
% \bibliography{ref}
% }

\bibliography{ref}

%%%%%%%%%%%%%%%%%%%%%%%%%%%%%%%%%%%%%%%%%%%%% Appendix %%%%%%%%%%%%%%%%%%%%%%%%%%%%%%%%%%%%%%%%%%%%%%%%%%%%%%
\newpage
% \bf{}
This is the arxiv version for the paper.
%%%%%%%%% Title
\begin{table*}[htbp]
    \begin{center}
         \resizebox{0.6\linewidth}{!}{
         \begin{tabular}{c}
            Supplementary Material (Appendix) \\
         \end{tabular}
         }
    \end{center}
    % \caption{Percentage of votes in the user study.}
    \label{table:title}
\end{table*}
\newpage
% \clearpage

\appendix

%%%%%%%%% BODY TEXT - ENTER YOUR RESPONSE BELOW
\section{Net Architecture}
{\flushleft \bf Encoder}.
The default dimension of the different stages in the encoder is 192, 384, and 768. The default strip width of the three stages is set to 2, 4, and 7. And the number of attention heads in the three stages is 3, 6, and 12.

{\flushleft \bf Transfer Module}.
The default dimension of the Transfer Module is 768 and the default number of attention heads is 8. And the default number of layers is three.

\begin{figure*}[htbp]
\centerline{\includegraphics[width=1.0\linewidth]{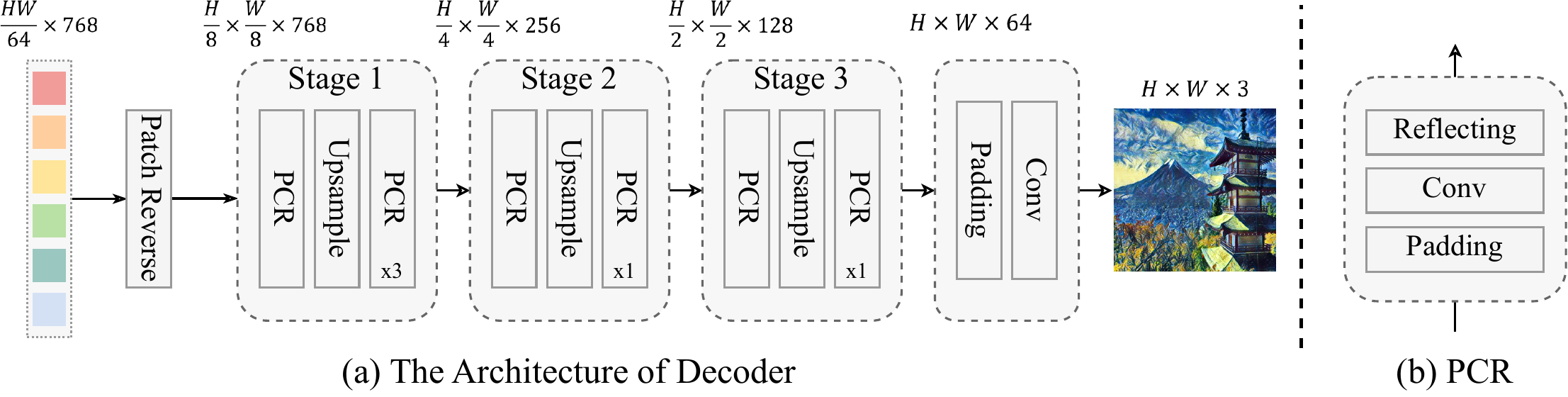}}
\caption{The architecture of decoder.}
\label{fig2a}
\end{figure*}

{\flushleft \bf Decoder}.
The decoder upscales the stylized features gradually, with the number of channels declining from 768 to 256 to 128 to 64, and finally downgrading the number of channels to 3. As shown in Fig.~\ref{fig2a}, given the stylized features in shape of $ \frac{WH}{64} \times C $, the patch reverse layer will reshape it to $ C \times \frac{H}{8} \times \frac{W}{8} $. Considering a $ Padding $, a $ Conv_{3\times3} $, and a $ ReLU $ as a basic processing unit (denotes as $ PCR $), the mirrored VGG upsamples the features in three stages. Stage one adopts a series of operations as $ PCR + Upsample_{\times 2} + 3 \times PCR $ while stages two and three take the same structure but subtract the last two $PCR$. Each stage upsamples features but subtracts channels. Finally, the features in shape of $ C^{'} \times H \times W $ will be processed by a $ Padding $ and a $ Conv_{3\times3} $ to reduce the channels to 3. Then comes the output image. 

\section{Differences from other methods}
{\flushleft \bf CSWin}.
In CSwin \cite{p90}, the strip-like window attention is used, while the fusion strategy is only concatenation which can not work in style transfer. Its shortcoming is described in the Section \textit{Experiments (First part)} of Appendix.
The necessity of the combination of multiple-shape window attentions and attn merge is also proved in the ablation study in main paper.
{\flushleft \bf Longformer and iLAT}.
\textit{First}, the fusion strategies for attention in Longformer \cite{p98} and iLAT \cite{p99} differ from our ``Attn Merge".
\textit{Second}, their attention computation manners are also varying from ours.
In Longformer, the global information is computed only from ``pre-selected input locations", while in SpW attention, each pixel can adaptively capture the information from both global (strip window) and local (square window) locations. 
Moreover, To achieve autoregressive generation, the attention of iLAT is limited in a fixed region (for the paper example 3x3+t[s]) and computes the strip-like global information with a mask.
In contrast, SpW does not constrain the fixed regions to calculate attention, and mask information is not required.

%------------------------------------------------------------------------

%-------------------------------------------------------------------------

%-------------------------------------------------------------------------
\section{Model Details in Pre-Analysis}
The models tested in Pre-Analysis of main paper adopt the encoder-transfer-decoder architecture, where the encoder is based on a Swin, which has 3 stages, and each stage has 2 successive Swin transformer blocks (we call it layer hereafter). When a layer is replaced by global MSA (multi-head self-attention), the window won't be shifted if it was. The transfer module and decoder are the same as those of S2WAT.

\section{Implementation Details}
\subsection{Dataset}
% {\flushleft \bf Dataset}.
% \noindent\textbf{Dataset.}
We use MS-COCO \cite{p84} and WikiArt \cite{p85} as the content dataset and the style dataset respectively. In the stage of training, the images are resized to 512 on the shorter side first and then randomly cropped to $ 224 \times 224 $. In the stage of testing, images of any size are accepted.

\subsection{Training Information}
% {\flushleft \bf Training Information}.
% \noindent\textbf{Training Information.} 
We use Pytorch \cite{p86} framework to implement S2WAT and train it for 40000 iterations with a batch size of 4. An Adam optimizer \cite{p87} and the warmup learning rate adjustment strategy \cite{p88} are applied with the initial learning rate of 1e-4. Our model is trained on a single Tesla V100 GPU for approximately 10 hours.

\section{Additional Experiments}
\subsection{Why do operations like sum and concatenation fail in the style transfer?}
The drawback of simple fusion operations, e.g., sum and concatenation, is that they will treat attention computation results from different windows equally. This characteristic may change the local content structure critically, thus hindering the optimization of content loss and leading to a monotonous style (as shown in Fig.~\ref{fig11} of the main paper).
E.g., assuming there are two features from different window attentions (one of them may focus on pixels of the sky and the other on the grassland), the outputs may not make sense when applying the sum or concatenation operation (a sky with grass?).

In contrast, our ``attn merge" can adaptively decide features to merge from different attentions. Thus, the structure of inputs can be reserved, and the local/global style information is consistent with the style reference image.

\subsection{Analysis for results of InST}
InST is a method of image style transfer based on Stable Diffusion Models (SDMs). InST mainly consists of three parts: the Textual Inversion module to transform style image inputs $I_s$ to the corresponding latent vectors $V_s$, the pre-trained SDM to denoise content noises $n_c$ to stylized images in condition of the latent vectors $V_s$, and the Stochastic Inversion module to predict noises from content inputs (the content noises $n_c$ are the results of adding the predict noises to the content images).

\begin{figure}[htbp]
\centerline{\includegraphics[width=1.0\linewidth]{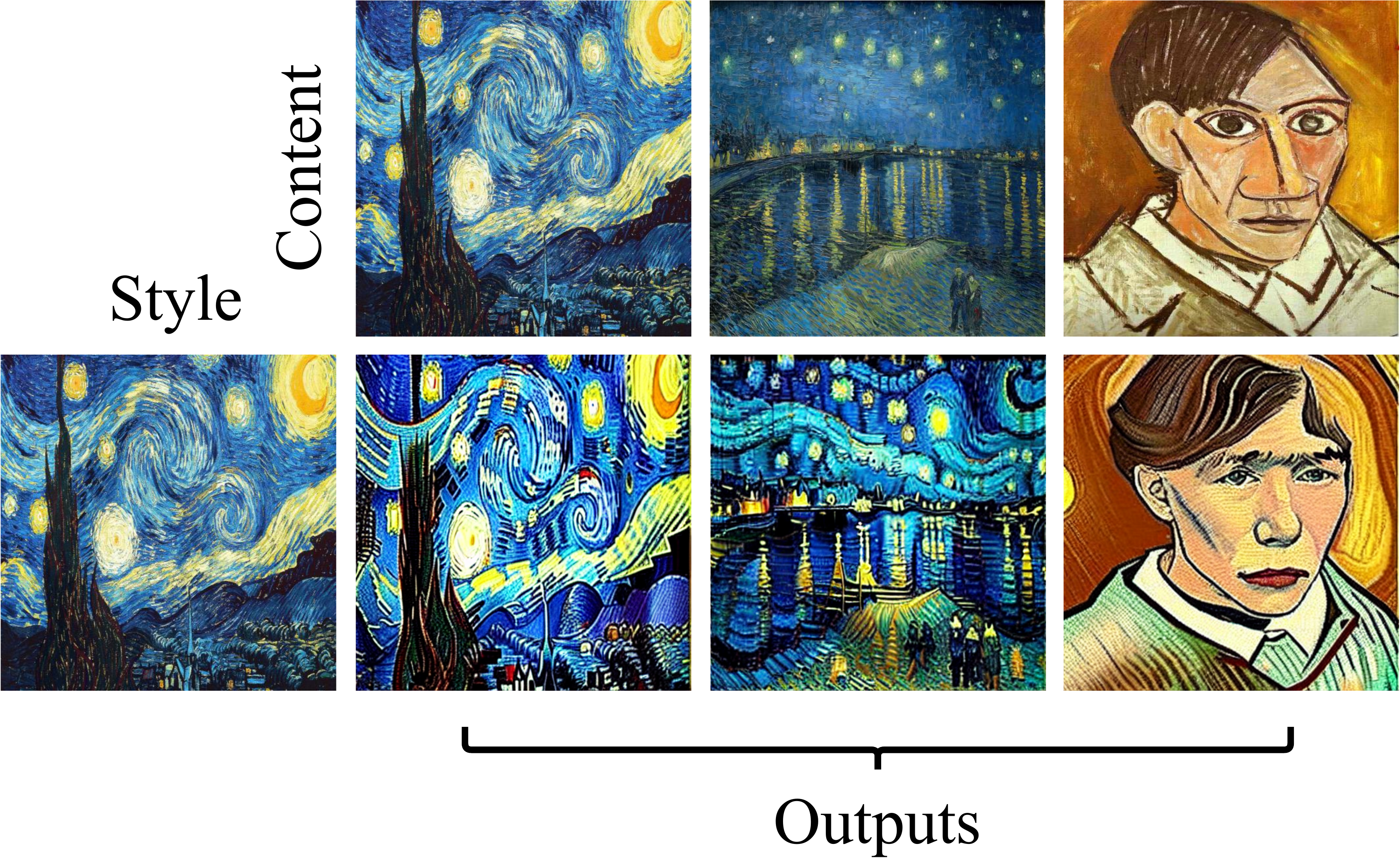}}
\caption{Results of InST in Different styles. The style similarity between the content images and the style images may affect the style transfer performance.}
\label{fig10a}
\end{figure}

During training, training data is a single style image (the content and style inputs are the same), which probably results in an overfitting problem. To avoid this problem, the dropout strategy is applied but it seems that the model still can not work well all the time. As presented in Fig.~\ref{fig10a}, InST performs normally when the style images are the same or similar to the content ones, while the results are unsatisfying if not. In the results of column 4, the style characteristics like colors fail to be transferred to the content image. Another possible reason may be the limited ability of the Textual Inversion module to encode specific style images, causing the loss of style information.

\subsection{Comparison with Swin-based Encoder}
In this section, we conduct experiments to compare results between the Swin-based model and S2WAT as part of the ablation study.

\vspace{1mm}
\noindent\textbf{Qualitative Comparison.} 
As depicted in Fig.~\ref{fig9a} (column 3), the results from the Swin-based model suffer from grid-like textures severely which can not be tolerated by users. For the results from the proposed S2WAT, this problem is fixed perfectly with smooth strokes and natural textures.

\noindent\textbf{Quantitative Comparison.} 
We also conduct a quantitative comparison for the Swin-based model as once done on the previous state-of-the-art methods. The results are presented in Table \ref{table:1a}. Surprisingly, S2WAT achieves the four best out of five metrics, which proves that S2WAT has delightful performance not only on the qualitative examination but also on the quantitative examination.

\begin{figure}[htbp]
\centerline{\includegraphics[width=1.0\linewidth]{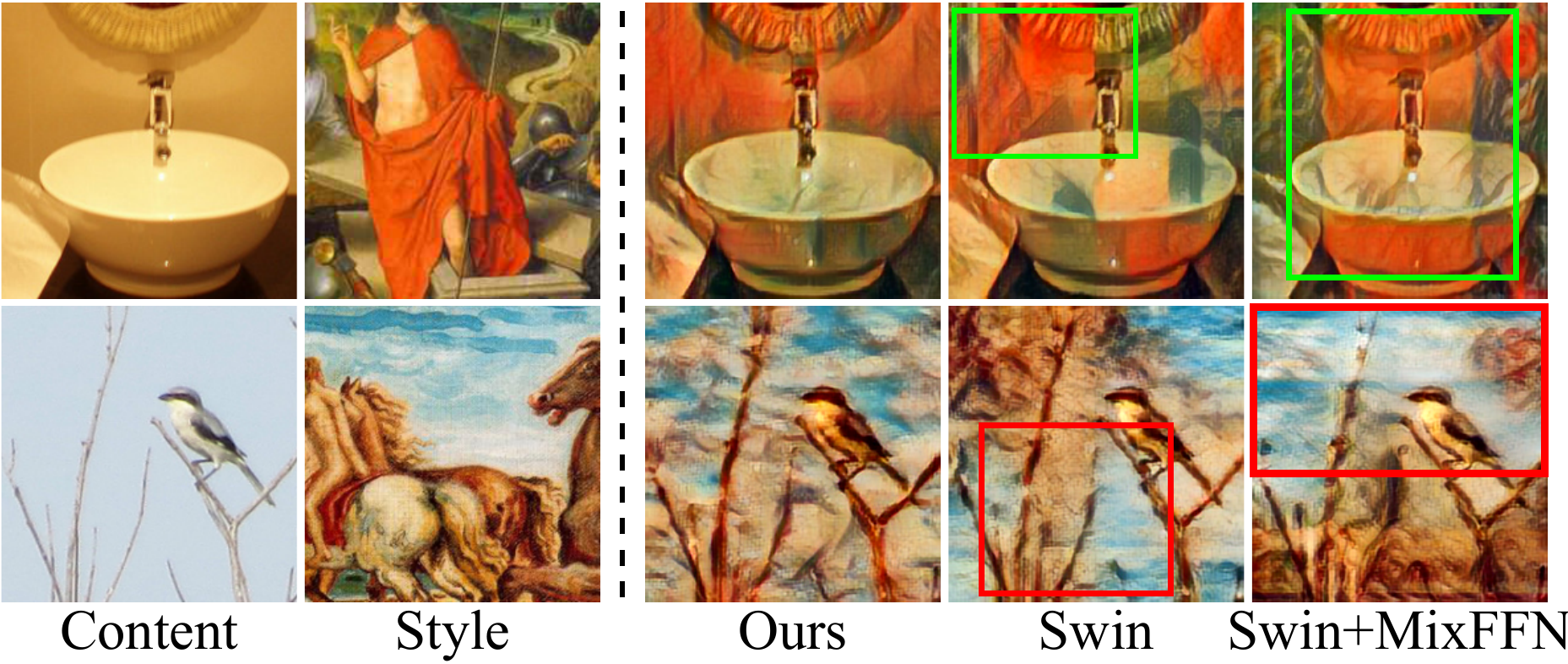}}
\caption{Results of combining Swin and Mix-FFN. Equipping Mix-FFN (mixture of convolutions and FFN) on Swin can not eliminate the blocky artifacts.}
\label{fig8a}
\end{figure}

\begin{figure}[htbp]
\centerline{\includegraphics[width=1.0\linewidth]{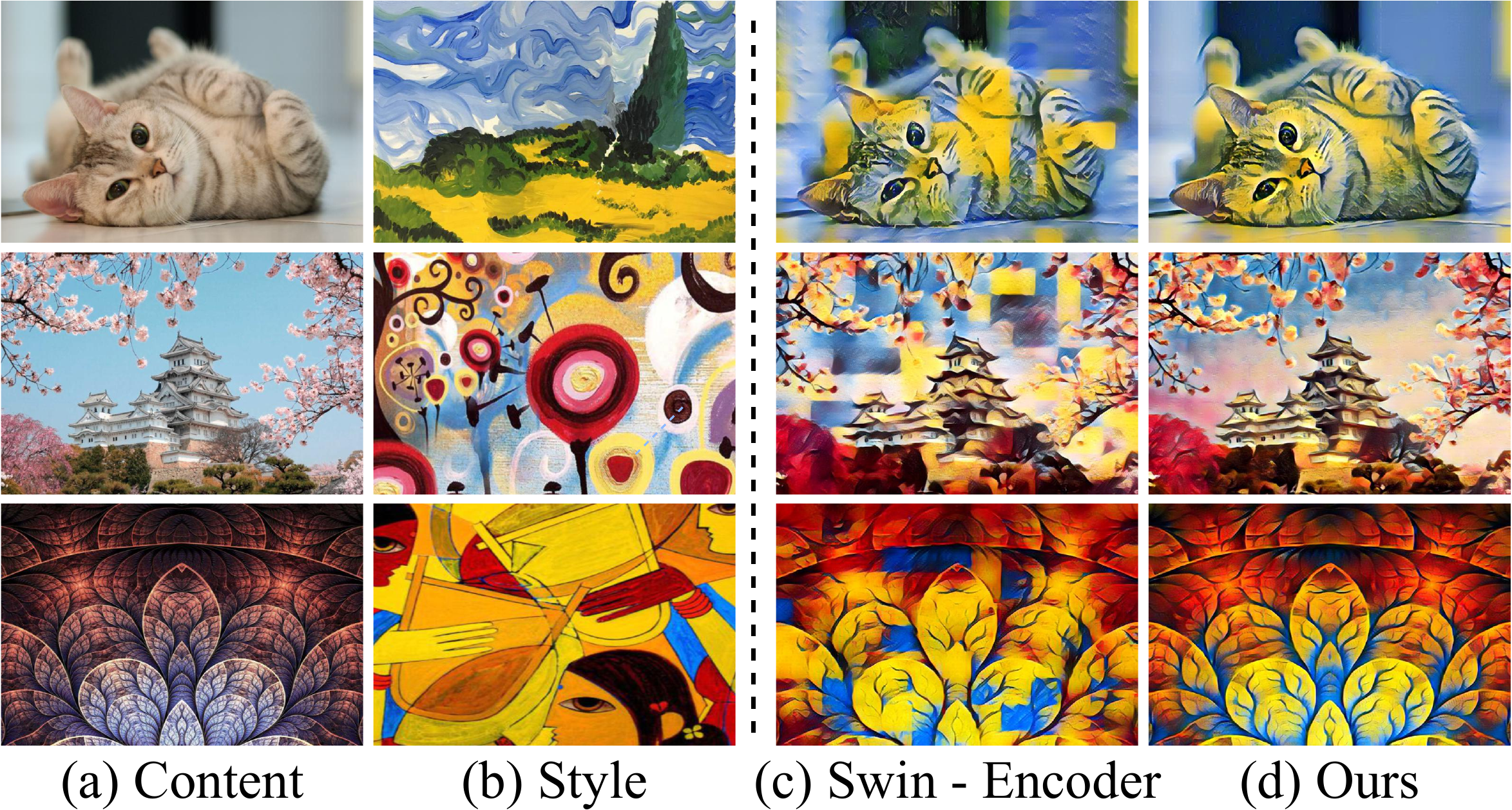}}
\caption{Ablation study for Swin-based encoder model. Obvious grid-like artifacts will appear when applying the Swin-based encoder.}
\label{fig9a}
\end{figure}

\begin{table*}[htbp]
    \begin{center}
         \resizebox{1.0\linewidth}{!}{
         \begin{tabular}{c c c c c c}
            \hline
            Model & Content Loss $\downarrow$ & Style Loss $\downarrow$ & Identity Loss $\downarrow$ 1 & Identity Loss $\downarrow$ 2 & SSIM $\uparrow$ \\
            \hline
            \textit{Swin-based} & 2.33 & 1.40 & 0.17 & 1.40 & 0.567 \\
            \textit{S2WAT} & \pmb{1.66} & 1.74 & \pmb{0.16} & \pmb{1.38} & \pmb{0.651} \\
            \hline
         \end{tabular}
         }
         % \vspace{-4mm}
    \end{center}
    \vspace{-0.1in}
    \caption{Quantitative evaluation results between the Swin-based model and S2WAT. The losses above are average values from 400 random samples while SSIMs are computed averagely from 100 samples. For a fair comparison, we take $ relu1\_1 $ into consideration in computing style loss and identity loss 2 while not in the training of S2WAT. The best results are in bold.}
    \label{table:1a}
    \vspace{-0.1in}
\end{table*}

\subsection{Comparison with Swin+Mix-FFN Encoder}
Mix-FFN \cite{p97} is a technique to make attention positional-encoding-free, which may be helpful to erase the grid-like texture. However, it seems that Mix-FFN can not work ideally in image style transfer. The results of combining Swin transformers with Mix-FFN are presented in Fig.~\ref{fig8a} (column 5). As shown in Fig.~\ref{fig8a}, blocky (or even strip-like) artifacts still appear on outputs, which proves that the local smooth of convolution can not erase the significant blocky artifacts.

%-------------------------------------------------------------------------

\subsection{Experiments with Attn Merge of only the Strip-based Attentions}
The model solely reliant on strip-based attentions (referred to as “Strips Only”) is insufficient due to two observed phenomena in our experiments: 1) Negative impact on quantitative metrics, especially the style loss values (see Table \ref{table:2a}); 2) Compromised generative quality, as demonstrated in Fig.~\ref{fig11a}.

\subsection{High Resolution Generation}
\begin{figure}[ht]
% \vspace{-0.1in}
\centerline{\includegraphics[width=1.0\linewidth]{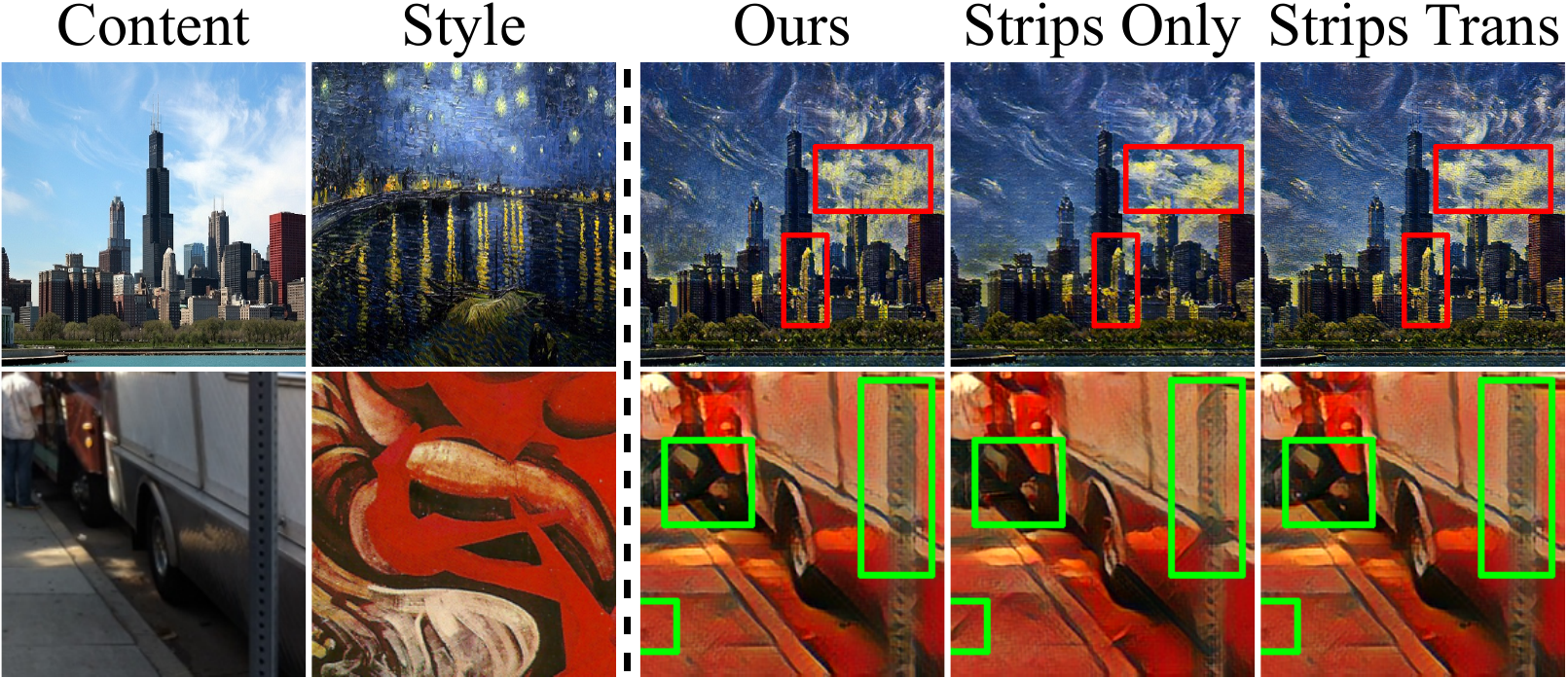}}
% \vspace{-0.05in}
\caption{Qualitative results between S2WAT and “Strips Trans/Only”. The red boxes highlight the impact on style, while the green boxes indicate undesirable strokes.}
\label{fig11a}
% \vspace{-0.15in}
\end{figure}

After completing training with a relatively larger strip width, S2WAT can afford higher resolution (up to $ 1024 \times 1024 $ on a 24G A5000 GPU) by reducing the strip width, with only a minor impact on performance. The first row of Table \ref{table:2a} and the last column of Fig.~\ref{fig11a} displays the results obtained by reducing the strip width of the 3-layer S2WAT from 2, 4, 7 to 1, 1, 1 (denoting as ``Strips Trans").

\begin{table}[t]
    \begin{center}
         \resizebox{1.0\linewidth}{!}{
            \begin{tabular}{c c c c c c}
                \hline
                Method & C-Loss $\downarrow$ & S-Loss $\downarrow$ & Id1 $\downarrow$ & Id2 $\downarrow$ & SSIM $\uparrow$ \\ 
                \hline
                Strips \, Trans & 1.69 & 1.85 & 0.16 & 1.39 & 0.647 \\
                Strips \, Only & 1.71 & 2.00 & 0.16 & 1.39 & 0.647 \\
                Ours           & \pmb{1.66} & \pmb{1.74} & \pmb{0.16} & \pmb{1.38} & \pmb{0.651} \\
                \hline
            \end{tabular}
         }
    \end{center}
    % \vspace{-4mm}
    \vspace{-0.1in}
    \caption{Quantitative results between S2WAT and “Strips Trans/Only”. C/S-Loss refers to content/style loss, while Id1/2 represents identity loss 1/2, respectively.}
    % \vspace{-0.2in}
    \vspace{-1.2em}
    \label{table:2a}
\end{table}

%-------------------------------------------------------------------------

\section{More Example of Style Transfer Results}
% The pictures presented in Fig.~\ref{fig3a}, Fig.~\ref{fig4a}, and Fig.~\ref{fig7a} are all in a resolution of $ 512 \times 768 \, (H \times W) $ and the ones in Fig.~\ref{fig5a} are in a resolution of $ 512 \times 512 $ .
The pictures presented in Fig.~\ref{fig3a} and Fig.~\ref{fig7a} are all in a resolution of $ 512 \times 768 \, (H \times W) $ and the ones in Fig.~\ref{fig5a} are in a resolution of $ 512 \times 512 $ .

\subsection{Qualitative Comparison}
\begin{figure*}[htbp]
\centerline{\includegraphics[width=1.0\linewidth]{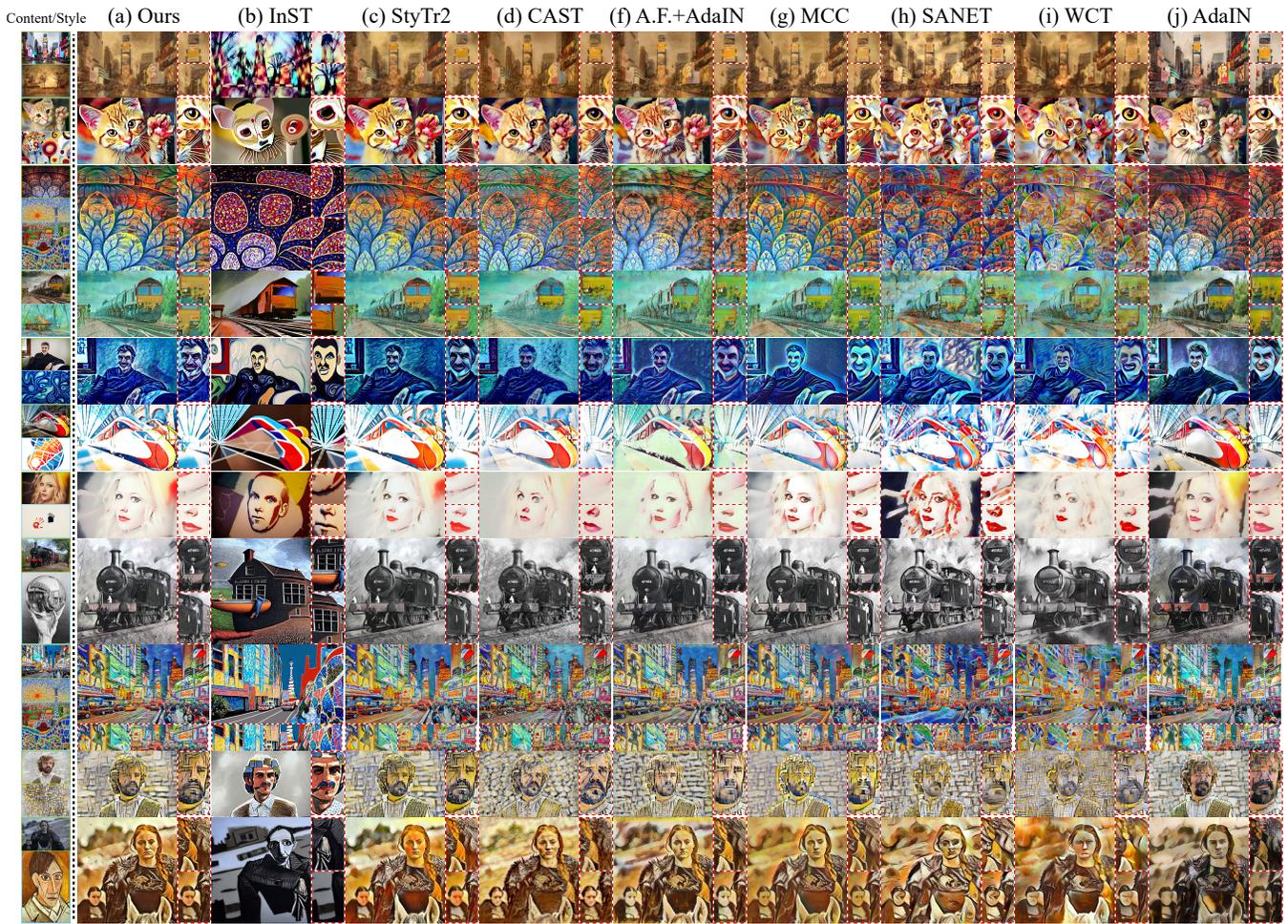}}
\caption{Extension of visual comparison between the results from state-of-the-art methods. A.F. denotes ``ArtFlow".}
\label{fig3a}
\end{figure*}

\begin{figure*}[htbp]
\centerline{\includegraphics[width=1.0\linewidth]{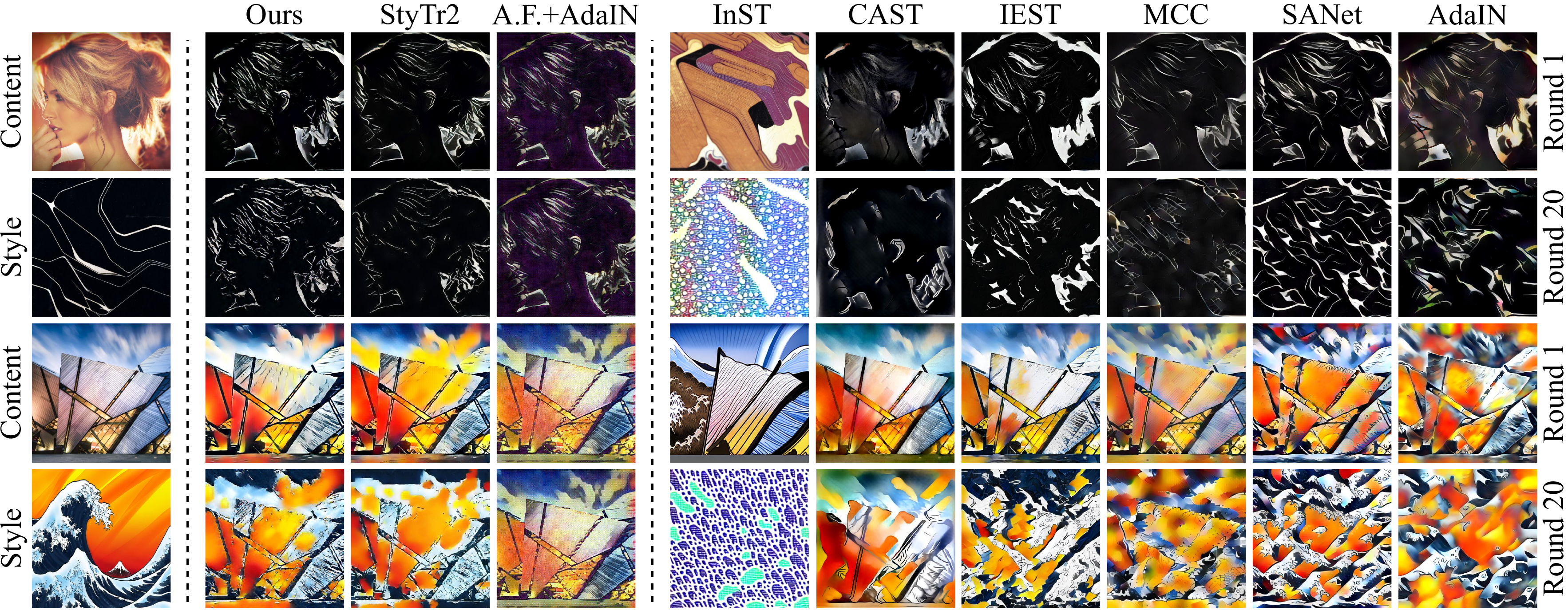}}
\caption{Extension of visualization on content leak. A.F. denotes ``ArtFlow".}
\label{fig5a}
\end{figure*}

\begin{figure*}[htbp]
\centerline{\includegraphics[width=1.0\linewidth]{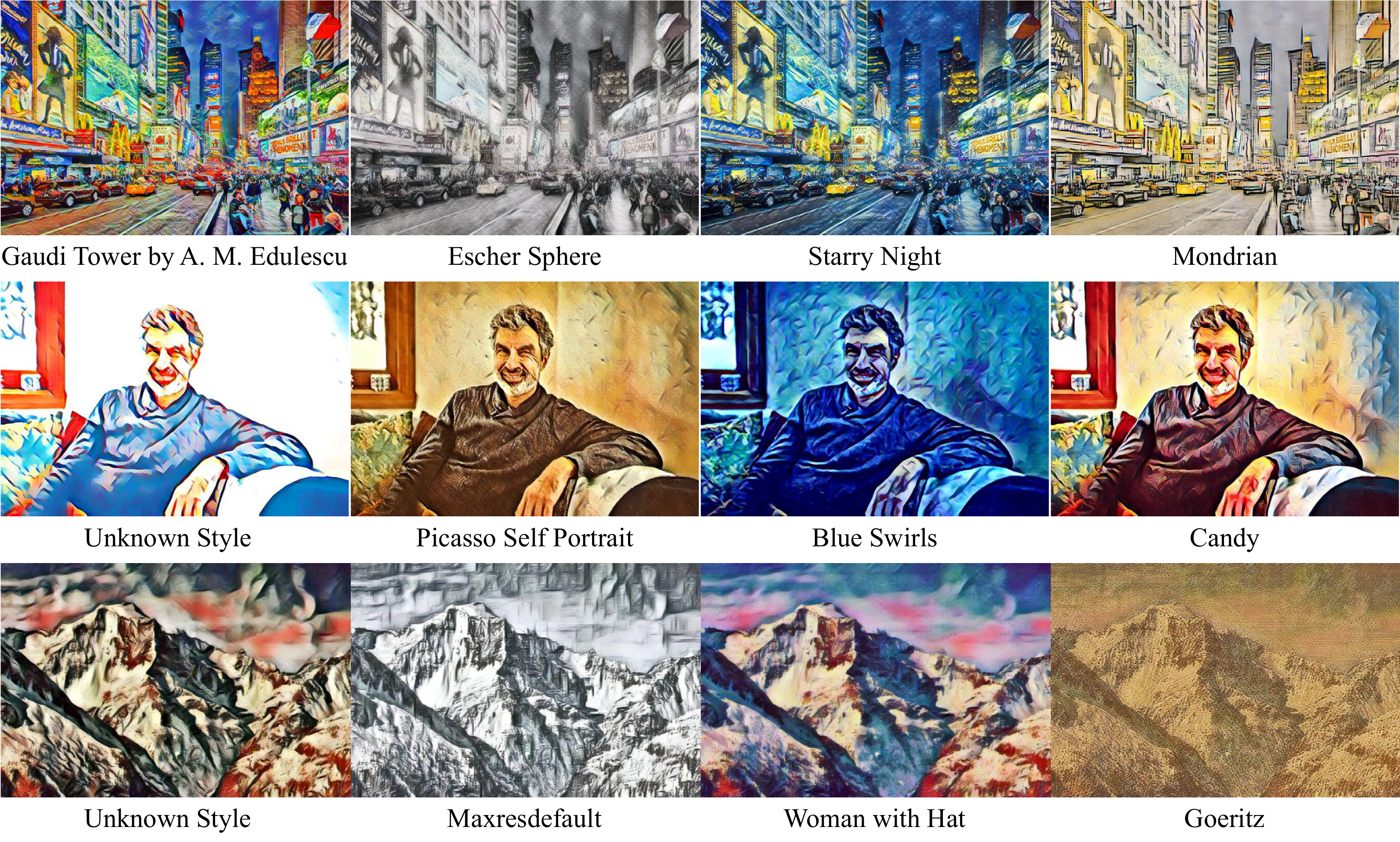}}
\caption{Results of multiple styles.}
\label{fig7a}
\end{figure*}

Due to the length limitation of 7 pages in the main body of the article, some of the results are removed in qualitative comparison (Section \textit{Style Transfer Results}). As shown in Fig.~\ref{fig3a}, we add an additional set of results in which the undesired patterns on the border from the results of ArtFlow \cite{p43} are more clear (rows 10 and 11 in Fig.~\ref{fig3a}). And more examples of the overflow issue from the results of MCC \cite{p41} could be found in row 5 and 10 of Fig.~\ref{fig3a}. To compare the style transfer ability between S2WAT and StyTr2, we also mark some areas of outputs with colored boxes which show that S2WAT is more capable of preserving content details.

% \begin{figure*}[htbp]
% \centerline{\includegraphics[width=1.0\linewidth]{figures/fig15_extension_of_visual_comparison.pdf}}
% \caption{Extension of visual comparison between the results from state-of-the-art methods.}
% \label{fig4a}
% \end{figure*}

%-------------------------------------------------------------------------

\subsection{Content Leak}
As shown in Fig.~\ref{fig5a}, an additional group of results under the content leak problem is provided (Section \textit{Content Leak}). Some content details though S2WAT will lose under a 20-round process, the results of S2WAT are obviously clear than that of CNN-based methods.

% \begin{figure*}[htbp]
% \centerline{\includegraphics[width=1.0\linewidth]{figures/fig16_extension_of_content_leak.pdf}}
% \caption{Extension of visualization on content leak.}
% \label{fig5a}
% \end{figure*}

\section{Details of User Study}
In conducting the user study, we leverage a widely-used online questionnaire platform employed by over 30 thousand corporations. A randomized selection of 31 content and style combinations, along with their corresponding stylized images, was presented to participants via a questionnaire. Before proceeding, participants were briefed on image style transfer and provided with evaluation criteria for an optimal outcome. These criteria encompassed two main aspects: a) a favorable outcome should diligently preserve content structure and details from the content image; b) a commendable outcome should effectively embody artistic attributes from the style image. Each participant was granted a 30-second timeframe to cast one or two votes, with all responses denoted in capital letters (omitting method names). After excluding outlier samples and instances where all votes were for a single method, we accumulated 3002 valid votes from 72 participants, including a diverse array of individuals ranging from high school students to professionals and experts spanning fields like computer science, art, and photography. The results of user study can be found in the Table~\ref{table:2} of the main paper.

We also provide an example questionnaire page we used in this part. As shown in Fig.~\ref{fig6a}, all the method names are covered with capitals as A, B, C, D, and E, where participants can not infer the corresponding relationship between results and methods.

\section{Limitation}
\begin{figure}[htbp]
\centerline{\includegraphics[width=1.0\linewidth]{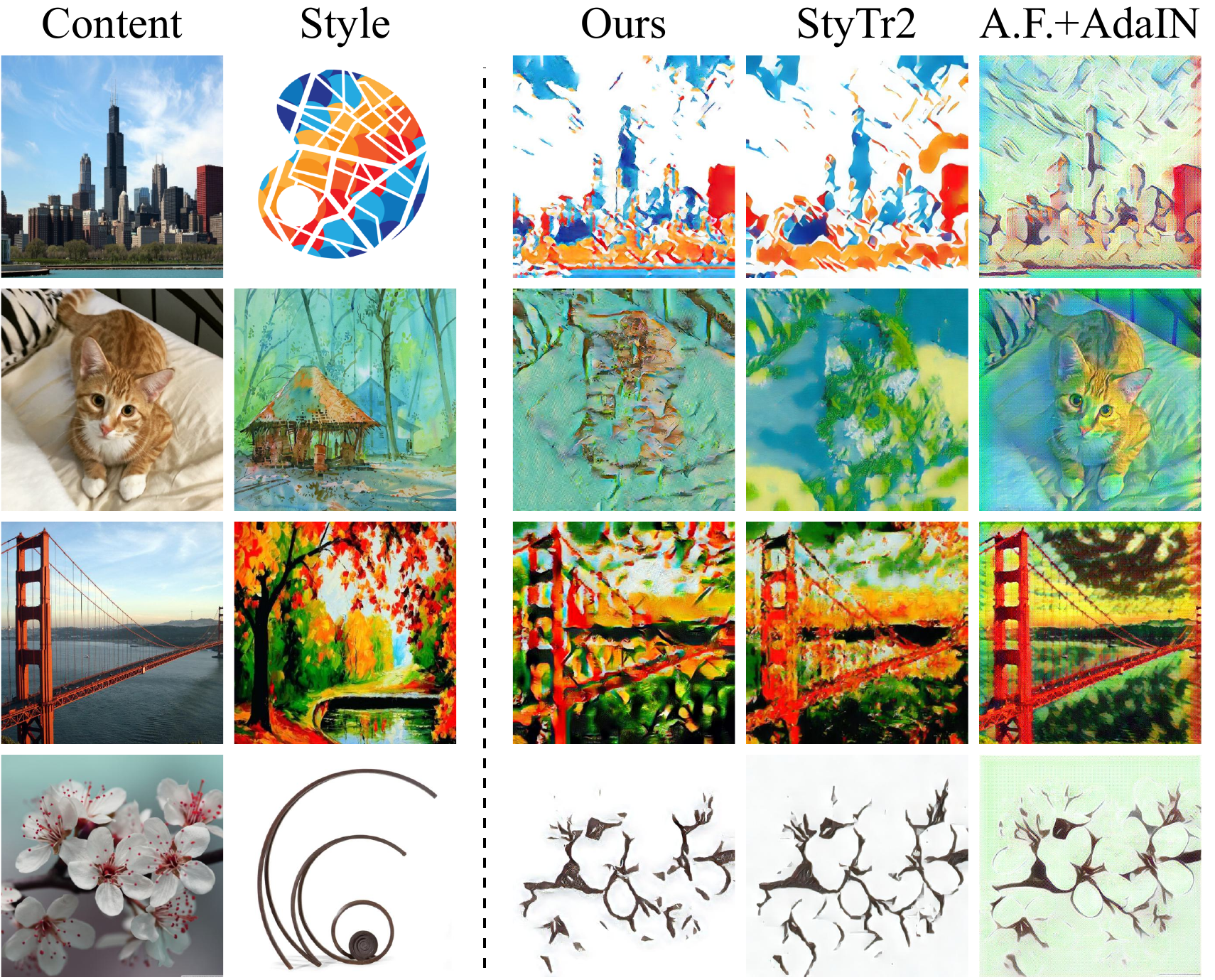}}
\caption{Limitation in content leak. The results above are all under a 20-round process. A.F. denotes ``ArtFlow".}
\label{fig1a}
\end{figure}

\begin{figure*}[htbp]
\centerline{\includegraphics[width=0.7\linewidth]{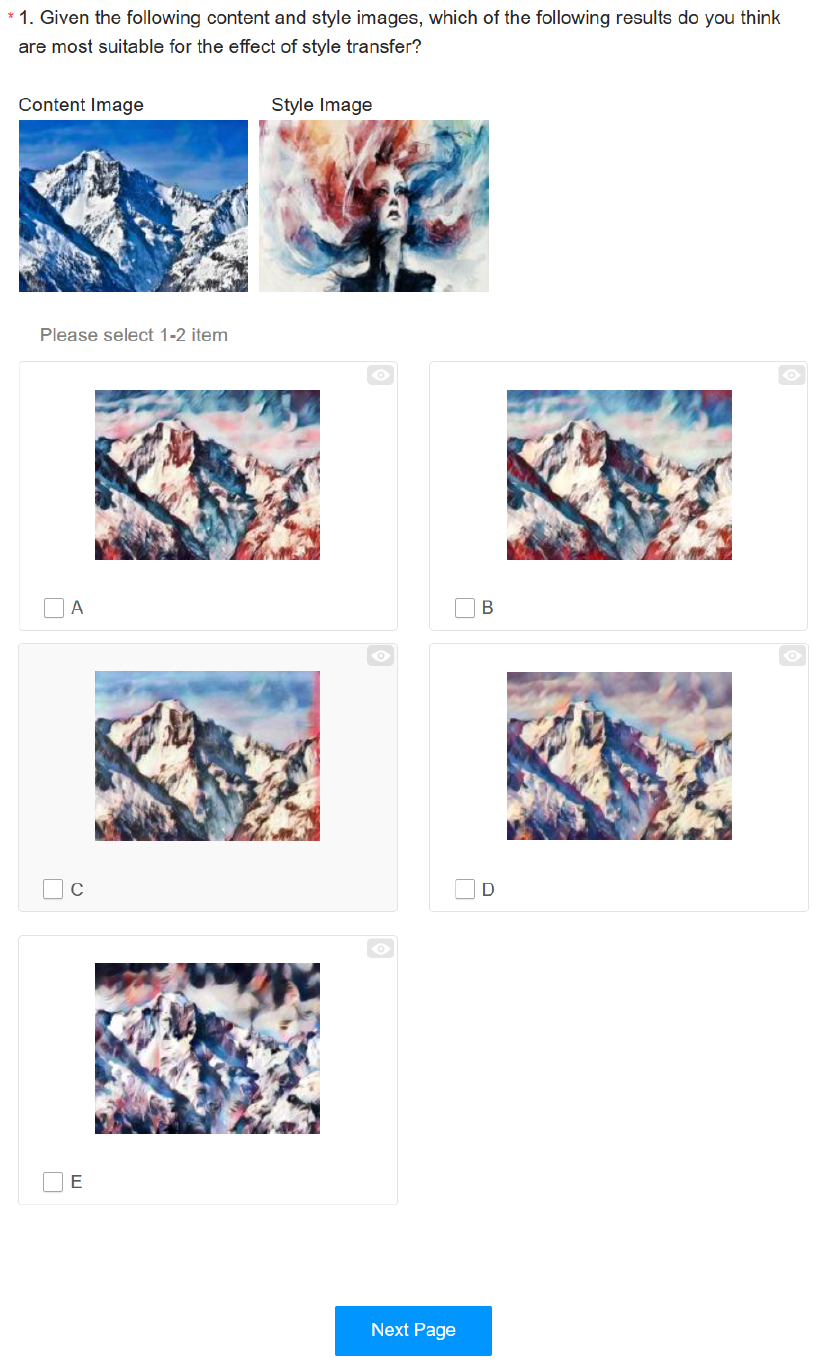}}
\caption{Example questionnaire page.}
\label{fig6a}
\end{figure*}

{\flushleft \bf Content Leak}.
As discussed on the performance under the content leak in Section \textit{Content Leak} of the main body of the article, S2WAT cannot achieves completely content-unbiased results. In some cases, the results may not be satisfying (as shown in Fig.~\ref{fig1a}). The reason why S2WAT is incompetent at producing completely content-unbiased results is that content details will be lost more or less from operations like downsampling. In the words of \cite{p43}, reconstruction errors will be accumulated by the biased modules or operations. However, compared to CNN-based methods, results from S2WAT have a clear advantage.

{\flushleft \bf Computational Complexity}.
Another limitation of S2WAT is the computational complexity of SpW Attention, which is also the next goal in our schedules. As discussed in Section \textit{Strips Window Attention} of the main body of the article, S2WAT cannot achieve linear computational complexity to the patch number. We try to replace the continuous sampling of window attention with discrete sampling to address this limitation in future work.

\end{document}